%% file: main.tex
\ifdefined\XeTeXversion\else
\pdfoutput=1
\fi
\documentclass{article}

\usepackage[T1]{fontenc}
\usepackage{microtype}
\usepackage{graphicx}
\usepackage{booktabs}
\usepackage{hyperref}

\usepackage[preprint]{icml2026}
\hypersetup{%
  pdftitle={GateAudit: Exact-Rational Risk-Limiting Audits with Causal Envelopes and Adaptive Re-Anchoring for Executable Agent Evaluation},%
  pdfauthor={Silu Panda},%
  pdfsubject={arXiv Preprint},%
  pdfkeywords={Risk-Limiting Audits, Anytime-Valid Inference, LLM Agents, Best-Arm Identification, Executable Evaluation},%
}

\usepackage{amsmath}
\usepackage{amssymb}
\usepackage{mathtools}
\usepackage{amsthm}
\usepackage{enumitem}

\theoremstyle{plain}
\newtheorem{theorem}{Theorem}
\newtheorem{proposition}{Proposition}
\newtheorem{lemma}{Lemma}
\newtheorem{corollary}{Corollary}
\theoremstyle{definition}

\theoremstyle{remark}
\newtheorem{remark}{Remark}

\newcommand{\E}{\mathbb{E}}
\newcommand{\Prob}{\mathbb{P}}
\newcommand{\Ind}{\mathbb{I}}
\newcommand{\Rcal}{\mathcal{R}}
\newcommand{\Fcal}{\mathcal{F}}
\newcommand{\Ucal}{\mathcal{U}}
\newcommand{\Kcal}{\mathcal{K}}
\newcommand{\Scal}{\mathcal{S}}

\input{generated_metrics.tex}
\input{generated_tables.tex}

\icmltitlerunning{GateAudit: Causal-Envelope Risk-Limiting Audits}

\begin{document}

\twocolumn[
  \icmltitle{GateAudit: Exact-Rational Risk-Limiting Audits with Causal Envelopes\\
    and Adaptive Re-Anchoring for Executable Agent Evaluation}

  \begin{icmlauthorlist}
    \icmlauthor{Silu Panda}{ind}
  \end{icmlauthorlist}

  \icmlaffiliation{ind}{Independent Researcher}
  \icmlcorrespondingauthor{Silu Panda}{\mbox{pandasilu428@gmail.com}}

  \icmlkeywords{Risk-Limiting Audits, Anytime-Valid Inference, LLM Agents, Best-Arm Identification, Executable Evaluation}

  \vskip 0.3in
]

\printAffiliationsAndNotice{}

\begin{abstract}
Selecting the best agent among $M$ candidates across $N$ executable tasks (\texttt{AppWorld}, \texttt{SWE-bench}, \texttt{EvalPlus}) nominally requires $NM$ stateful rollouts. Fixed-budget evaluators (\texttt{PULSE}, \texttt{UCB-E}, Sequential Halving) lack anytime $\delta$-stopping guarantees, fixed-anchor risk-limiting audits (RLAs) lose efficiency when proxies invert rankings, and confidence-sequence bandits (\texttt{WOR-CS LUCB}) pay an $M$-way union bound while ignoring execution causality. We introduce \textbf{GateAudit}, a finite-sample, exact-rational framework certifying the exact tie-broken winner of a realized envelope-admissible matrix at risk $\delta \in (0,1)$ via: (i) \textbf{\texttt{CPPS-Mix}} (Theorem~\ref{thm:cpps}), an exact-rational 4-point mixture supermartingale over Lazy-Row and Atomic-Cell units with zero $(M{-}1)$ union bound; (ii) engine-agnostic \textbf{causal envelope tightening} (Theorem~\ref{thm:envelope}) for prefix-invariant step-$K$ halting (\texttt{CPPS-HC}) and subset tests (\texttt{CPPS-OC}), paired with a frequentist hypergeometric \textbf{Envelope Spot-Check} (Corollary~\ref{cor:env_spot}); and (iii) \textbf{\texttt{GateAudit-Unified}} (Theorem~\ref{thm:unified}), combining multiplicity-free Phase~A ($\delta_1{=}4\delta/5$) with conditional Phase~B survivor re-anchoring ($\delta_2{=}\delta/5$). Across $40$ real \texttt{SWE-bench Verified} runs (counterfactual truncation), \texttt{EvalPlus}, and live \texttt{AppWorld} evaluations on three GPUs (\texttt{Qwen3-8B} plus two zero-signal \texttt{Llama-3.1} tie-break runs), GateAudit cuts oracle/continuation cells $54\%\text{--}87\%$ at $K{=}40$/\texttt{EvalPlus} ($6.6\%\text{--}83.6\%$ across $K{\in}\{10..40\}$; $1.33{\times}\text{--}3.60{\times}$ fewer than \texttt{Native PULSE CI-stop+Env} at $M{=}12$, $0.99{\times}\text{--}1.66{\times}$ at $M{=}4$), caps worst-case empirical regret vs.\ $\min(\texttt{CPPS-Mix}, \texttt{WOR-CS+Env})$ at $8.8\%$, and cuts live GPU wall-clock $19.8\%$ vs.\ Stage-0 \texttt{H100} \texttt{E1} ($\approx 21\%\text{--}24\%$ order-adjusted on \texttt{A100}) while respecting $\delta{=}5\%$ on all well-specified deterministic envelopes.
\end{abstract}

\section{Introduction}
\label{sec:intro}

Developing tool-using language agents \citep{yao2023react,kang2026acon} and code-generation systems \citep{jimenez2024swebench,liu2023evalplus} relies on repeated \textbf{fixed-suite winner selection}: given $M$ candidate harnesses or model checkpoints and $N$ executable tasks, identify the exact tie-broken winner on the realized pass/fail matrix $S^* \in \{0, 1\}^{N \times M}$. Exhaustively running all $NM$ trajectories is prohibitive when each cell entails tens of reasoning turns against stateful SQLite databases \citep{trivedi2024appworld}, Docker repositories \citep{jimenez2024swebench}, or property-based test suites \citep{liu2023evalplus}.

Recent evaluation accelerators use prediction-powered inference \citep{angelopoulos2023ppi,boyeau2024autoeval,zrnic2024active}, anchor subsets \citep{polo2024tinybenchmarks,vivek2024anchor}, LLM-judge routing \citep{ao2026service}, or low-rank matrix completion (\texttt{PULSE} \citep{pulse2026active}). Yet executable agent evaluation exposes four structural bottlenecks:
\begin{enumerate}[leftmargin=1.4em,itemsep=0.5pt,topsep=1pt]
  \item \textbf{Non-causal inversion vs.\ causal prefix invariance.} Disabling \texttt{<think>} on live \texttt{H100} \texttt{AppWorld} (\texttt{Qwen3-8B} \citep{yang2025qwen3}) alters Turn-1 tokens and flips $11/13$ ($84.6\%$) of task-level candidate-pair outcomes where $s^*_{ic_1} \neq s^*_{ic_2}$ ($[4,3,2,2]\!\to\![1,3,3,2]$). Conversely, truncating \emph{only} \texttt{max\_steps}$=K$ under deterministic decoding preserves prefix steps $1\dots K$: halted runs ($H_{ic}=1$) settle $s^*_{ic} = \hat{s}_{ic}$ exactly, while unhalted runs gated to $\hat{s}_{ic}=0$ satisfy $s^*_{ic} \ge \hat{s}_{ic}$.
  \item \textbf{Two-sided slack vs.\ causal envelopes.} Classical comparison RLAs \citep{stark2020shangrla,stark2023alpha} charge discrepancy caps on already-settled cells, consuming $84\%\text{--}99\%$ of $NM$ on tight \texttt{SWE-bench} margins. One-sided causal envelopes cut oracle cells roughly in half, though silent violations on skipped cells require explicit violation-rate spot-checking.
  \item \textbf{Fixed-anchor inversion vs.\ $M$-way union bounds.} Fixed-anchor RLAs avoid an $(M{-}1)$ union bound---beating confidence-sequence bandits (\texttt{WOR-CS LUCB}) at large $M$ when $\hat{w} = w^\star$---but lose efficiency when proxies invert rankings ($\hat{w} \neq w^\star$).
  \item \textbf{Oracle cells vs.\ serving units and API dollars.} Cell counts ignore task fixture/prefill setup $p$ \citep{kwon2023vllm,zheng2024sglang} and upfront prefix spend: \textbf{Atomic-Cell} wins at $p/m \ll 1$, \textbf{Lazy-Row} wins at $p/m \gg 1$, and unsunk frontier API prefixes can cost more USD than proxy-free \texttt{WOR-CS LUCB} without Stage-0 cost routing.
\end{enumerate}

\begin{figure*}[t]
  \centering
  \includegraphics[width=0.90\textwidth]{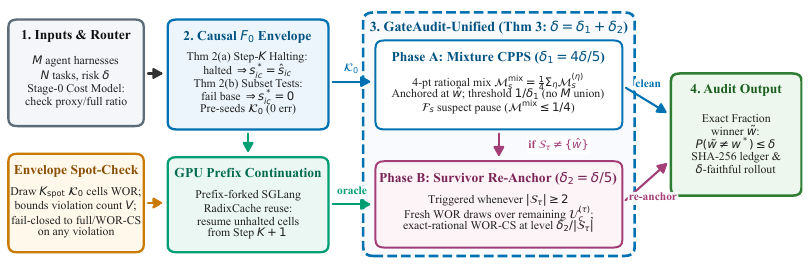}
  \vspace{-0.9em}
  \caption{\textbf{End-to-end GateAudit architecture.} Stage-0 prices proxy vs.\ direct full evaluation; an optional hypergeometric spot-check tests the violation count $V$ of $\Kcal_0$ (Corollary~\ref{cor:env_spot}); GPU continuation resumes unhalted cells from Step $K{+}1$ over warm \texttt{RadixCache}; and \texttt{GateAudit-Unified} (Theorem~\ref{thm:unified}) pairs multiplicity-free 4-point mixture \texttt{CPPS-WOR} ($\delta_1 = 4\delta/5$) with conditional survivor re-anchoring ($\delta_2 = \delta/5$) in exact \texttt{Fraction} arithmetic.}
  \label{fig:arch}
  \vspace{-0.7em}
\end{figure*}

\paragraph{Contributions.} \textbf{GateAudit} (Figure~\ref{fig:arch}; code at \url{https://github.com/SiluPanda/gateaudit}) addresses all four bottlenecks in a pip-installable, exact-rational package:
\begin{itemize}[leftmargin=1.3em,itemsep=0.5pt,topsep=1pt]
  \item \textbf{Causal Envelopes \& Spot-Check (Theorem~\ref{thm:envelope}, Corollary~\ref{cor:env_spot}):} One-sided caps for step-$K$ halting (\texttt{CPPS-HC}) and subset tests (\texttt{CPPS-OC}) pre-seed settled cells $\Kcal_0$ across any engine (\texttt{CPPS}, \texttt{WOR-CS+Env}), paired with a frequentist hypergeometric violation-rate audit ($K_{\mathrm{spot}}{=}150$ detects $94.0\%$ of $2\%$-violation runs, cutting error from $31.5\%$ to $0.0\%$).
  \item \textbf{Exact-Rational \texttt{CPPS-Mix} \& \texttt{Unified} (Theorems~\ref{thm:cpps},~\ref{thm:unified}):} A 4-point rational betting mixture plus adaptive Phase~A/B survivor re-anchoring. Across all $16$ \texttt{SWE-bench} $M$-configs, \texttt{Unified} caps worst-case empirical regret vs.\ $\min(\texttt{CPPS-Mix}, \texttt{WOR-CS+Env})$ at $8.8\%$ (vs.\ $18.8\%$ \texttt{WOR-CS+Env}, $31.1\%$ \texttt{CPPS-Mix}), while \texttt{CPPS-Mix} saves $-15.8\%$ (\texttt{Unified} $-8.4\%$) vs.\ \texttt{WOR-CS+Env} on clean $M{=}30$.
  \item \textbf{Stochastic Faithfulness \& Systems (Proposition~\ref{prop:stochastic}, \S\ref{sec:systems}):} Under batch-invariant seeding, auditing an $NR \times M$ replicate matrix adds $\le \delta_{\mathrm{audit}}$ to full-evaluation error; Turn-1 forking and Step-$K$ continuation unlock live GPU speedups.
  \item \textbf{Honest Multi-Benchmark \& Multi-GPU Evaluation:} Envelope-seeded \texttt{+Env} baselines (\texttt{Native PULSE CI-stop+Env}, \texttt{SH+Env}, \texttt{UCB-E+Env}), $40$-model \texttt{SWE-bench Verified} (\texttt{Rand-10}, \texttt{Frontier}, \texttt{Clean}), USD vs.\ ms accounting, and five live \texttt{AppWorld} GPU runs (\texttt{H100}, \texttt{Blackwell}, \texttt{A100}; \texttt{Qwen3-8B}, \texttt{Llama-3.1}).
\end{itemize}

\section{Problem Formulation and Physical Cost Model}
\label{sec:setup}

\paragraph{Fixed-suite tie-broken winner.} Consider $N$ tasks $i \in [N] \coloneqq \{1,\dots,N\}$ with integer weights $w_i \in \mathbb{Z}_{\ge 1}$ (default $1$) and $M \ge 2$ frozen candidates $c \in [M]$ with strict tie-break ranks $r_c \in \{0,\dots,M{-}1\}$ (smaller wins ties). Full evaluation yields binary outcomes $s^*_{ic} \in \{0,1\}$, totals $T^*_c \coloneqq \sum_{i=1}^N w_i s^*_{ic}$, and unique exact winner
\begin{equation}
  w^\star \coloneqq \arg\min_{c \in [M]} \bigl(-T^*_c,\, r_c\bigr).
  \label{eq:winner}
\end{equation}
Optionally, a provisional binary proxy matrix $\hat{S} \in \{0,1\}^{N \times M}$ (with halting mask $H \in \{0,1\}^{N \times M}$ under \texttt{CPPS-HC}) gives totals $\hat{T}_c \coloneqq \sum_{i=1}^N w_i \hat{s}_{ic}$ and provisional leader $\hat{w} \coloneqq \arg\min_c(-\hat{T}_c, r_c)$.

\paragraph{Risk-limiting certification guarantee.} Fix an admissible class $\mathcal{C}$ of deterministic instances $(\hat{S}, H, S^*)$ ($\mathcal{C}$ is unconstrained under two-sided mode, and restricted to envelope-consistent instances under Theorem~\ref{thm:envelope}). Using fresh internal randomness independent of $S^*$, an auditor queries oracle units and outputs $\tilde{w} \in [M]$. Given $\delta \in (0, 1)$, we require
\begin{equation}
  \Prob_{\mathrm{audit}}\bigl(\tilde{w} \neq w^\star \;\big|\; \hat{S}, H, S^*\bigr) \;\le\; \delta, \qquad \forall (\hat{S}, H, S^*) \in \mathcal{C},
  \label{eq:rla}
\end{equation}
including adversarial ties and inverted proxies ($\hat{w} \neq w^\star$).

\paragraph{Purchase units and logical wave filtration.} Oracle queries map to \textbf{Lazy-Row} ($\Ucal = [N]$, opening task $i$'s fixture once for active candidates---and only unhalted cells under \texttt{HC}) or \textbf{Atomic-Cell} ($\Ucal = [N] \times [M]$, executing rollout $(i, c)$ alone). Each parallel wave pre-commits an ordered sequence $(I_{t+1},\dots,I_{t+k})$ ($k \le b_{\mathrm{cells}}$) via without-replacement sampling; workers execute asynchronously while wealth updates process $(I_{t+1},\dots,I_{t+k})$ strictly in logical draw order, keeping $\Fcal_{s-1} \coloneqq \sigma(I_1,\dots,I_{s-1})$ invariant to latencies.

\begin{proposition}[Admissible Pivotal-Set Task Lower Bound]
\label{prop:lower}
Let $(\hat{S}, H, S^*) \in \mathcal{C}$ be a two-candidate instance where $A$ beats $B$, and let $\Kcal \subseteq [N]$ be the admissible pivotal task set---each $i \in \Kcal$ admitting $(\hat{S}, H, \tilde{S}^{(i)}) \in \mathcal{C}$ agreeing with $S^*$ on $[N]\setminus\{i\}$ on which $B$ beats $A$. Any black-box audit satisfying~\eqref{eq:rla} at $\delta < 1/2$ opens expected distinct tasks $\E[d \mid \hat{S}, H, S^*] \ge (1 - 2\delta)|\Kcal|$.
\end{proposition}

\begin{remark}[Physical Occupancy Crossover]
\label{rem:reversal}
Under expected additive work $\E[T] = \E[d]\cdot p + \E[n_{\mathrm{c}}]\cdot m + \E[n_{\mathrm{w}}]\cdot q$ (setup $p$, rollout $m$, wave barrier $q$, with unit counts fixed across $p/m$), let $\Delta n_{\mathrm{c}} \coloneqq \E[n_{\mathrm{c,row}}] - \E[n_{\mathrm{c,atom}}]$ and $\Delta n_{\mathrm{w}} \coloneqq \E[n_{\mathrm{w,row}}] - \E[n_{\mathrm{w,atom}}]$. Whenever $\E[d_{\mathrm{atom}}] > \E[d_{\mathrm{row}}]$, Lazy-Row beats Atomic-Cell iff
\[
  \frac{p}{m} \;>\; \Bigl(\frac{p}{m}\Bigr)^{\!\star} \;\coloneqq\; \frac{\Delta n_{\mathrm{c}} + \frac{q}{m}\,\Delta n_{\mathrm{w}}}{\E[d_{\mathrm{atom}}] - \E[d_{\mathrm{row}}]}.
\]
\end{remark}

\begin{remark}[Common-Offset Cancellation]
\label{rem:common_offset}
When proxy errors share a task-level shock ($\hat{s}_{ic} - s^*_{ic} = \xi_i$), Lazy-Row cancels $\xi_i$ identically inside $e_i(a)=0$, whereas unpaired Atomic-Cell draws on $(i,\hat{w})$ yield $B_{i\hat{w}}(a) = w_i s^*_{i\hat{w}}$ (dropping to $0$ when $\xi_i=+1, s^*_{i\hat{w}}=0$) until paired draws cancel the offset.
\end{remark}

\paragraph{Composition with stochastic rollouts.} For stochastic decoding ($\mathrm{temp} > 0$) on a fixed $N$-task suite with per-arm expected pass rate $\mu_c \coloneqq \frac{1}{N}\sum_{i=1}^N \E[s^*_{ic}]$ over rollout randomness, drawing $R \ge 1$ i.i.d.\ rollouts per $(i, c)$ defines an $NR \times M$ realized matrix $S^*$ with full-evaluation winner $w^\star(S^*)$.

\begin{proposition}[$\delta$-Faithfulness under Batch-Invariant Stochastic Rollouts]
\label{prop:stochastic}
Assume each cell's stochastic outcome $s^*_{(i,r),c}$ is a deterministic, batch-invariant function of an independent pre-assigned RNG seed $\xi_{irc}$ independent of the auditor's sampling stream (deferred-decision coupling). Auditing the realized $NR \times M$ matrix on demand via an engine satisfying~\eqref{eq:rla} at risk $\delta_{\mathrm{audit}}$ guarantees, for $w_{\mathrm{pop}}^\star = \arg\min_c(-\mu_c, r_c)$,
\[
  \Prob\bigl(\tilde{w} \neq w_{\mathrm{pop}}^\star\bigr) \;\le\; \Prob\bigl(w^\star(S^*) \neq w_{\mathrm{pop}}^\star\bigr) + \delta_{\mathrm{audit}}.
\]
\end{proposition}
Proofs appear in Appendix~\ref{app:proofs_props}. Note that Proposition~\ref{prop:stochastic} requires within-stack batch-invariant inference---empirically consistent on our \texttt{H100} \texttt{FA3} run (logged with \texttt{--attention-backend fa3 --enable-deterministic-inference}; $0/58$ halted, $12/12$ continuation, and $72/72$ revealed matches), whereas high-batch \texttt{sm\_80} Triton showed within-stack drift (\S\ref{sec:exp_live}).

\section{Exact-Rational Mixture Audits, Causal Envelopes, and Unified Re-Anchoring}
\label{sec:theory}

\subsection{Theorem 1: Exact-Rational Mixture \texttt{CPPS-WOR}}
\label{sec:thm1}

For each rival $a \in [M] \setminus \{\hat{w}\}$, let $m_a \coloneqq \hat{T}_{\hat{w}} - \hat{T}_a$ and tie-aware threshold $h_a \coloneqq m_a + \Ind_{r_a > r_{\hat{w}}}$. Rival $a$ outranks $\hat{w}$ on $S^*$ iff total discrepancy $E_a \coloneqq \sum_{u \in \Ucal} e_u(a) \ge h_a$, where for Lazy-Row ($\Ucal=[N]$) and Atomic-Cell ($\Ucal=[N]\times[M]$):
\begin{align}
  e_i(a) &\coloneqq w_i\bigl[(\hat{s}_{i\hat{w}} - \hat{s}_{ia}) - (s^*_{i\hat{w}} - s^*_{ia})\bigr],
  \nonumber\\[-2pt]
  U_i(a) &\coloneqq w_i \bigl(1 + \hat{s}_{i\hat{w}} - \hat{s}_{ia}\bigr), \label{eq:row_cap}\\
  e_{ic}(a) &\coloneqq w_i \Ind_{c=\hat{w}}(\hat{s}_{i\hat{w}}{-}s^*_{i\hat{w}}) - w_i \Ind_{c=a}(\hat{s}_{ia}{-}s^*_{ia}),
  \nonumber\\[-2pt]
  U_{ic}(a) &\coloneqq w_i \hat{s}_{i\hat{w}}\Ind_{c=\hat{w}} + w_i(1{-}\hat{s}_{ia})\Ind_{c=a}. \label{eq:cell_cap}
\end{align}
Pointwise, $e_u(a) \le U_u(a)$, so $B_u(a) \coloneqq U_u(a) - e_u(a) \ge 0$, and $U_u(a)=0 \implies e_u(a) \le 0$.

\paragraph{Active-conditioned PPS-WOR mixture betting.} Let $\mathcal{O}_{s-1}$ denote all previously revealed units (including any $\Fcal_0$-settled units $\Kcal_0$ under Theorem~\ref{thm:envelope}), $P_{s-1} = \Ucal \setminus \mathcal{O}_{s-1}$, and $\Rcal \subseteq [M]\setminus\{\hat{w}\}$ the active rival set. At wave start, freeze PPEB-style integer weights $W_u \coloneqq \max_{b \in \Rcal} U_u(b)$ \citep{aslam2008ppeb} and draw $(I_{t+1},\dots,I_{t+k})$ sequentially without replacement from $P^+_{s-1} \coloneqq \{u \in P_{s-1} : W_u > 0\}$ with step probability $W_u / \sum_{v \in P^+_{s-1}} W_v$ via exact integer-ticket \texttt{randrange}. Before step $s$, let $A_{s-1}(a) \coloneqq \{u \in P_{s-1} : U_u(a) > 0\}$, $W_{\mathrm{act}}(a) \coloneqq \sum_{u \in A_{s-1}(a)} W_u$, $U_{\mathrm{rem}}(a) \coloneqq \sum_{u \in A_{s-1}(a)} U_u(a)$, $r_{s-1}(a) \coloneqq h_a - \sum_{u \in \mathcal{O}_{s-1}} e_u(a)$, and slack $D_{s-1}(a) \coloneqq U_{\mathrm{rem}}(a) - r_{s-1}(a)$.

Fixed-$\eta$ betting ($\eta{=}17/20$) grows quickly on large margins but loses wealth rapidly on noisy, small-margin discrepancies. We instead maintain an \textbf{exact-rational uniform discrete mixture} over $\mathcal{E} = \{\frac{1}{4}, \frac{1}{2}, \frac{3}{4}, \frac{17}{20}\}$: each component $\mathcal{M}_s^{(\eta)}(a) = \mathcal{M}_{s-1}^{(\eta)}(a) F_s^{(\eta)}(a)$ ($\mathcal{M}_0^{(\eta)}(a)=1$) sets $F_s^{(\eta)}(a) \coloneqq 1$ if $I_s \notin A_{s-1}(a)$ or $(D_{s-1}(a){=}0 \land B_{I_s}(a){=}0)$, and otherwise (when $I_s \in A_{s-1}(a)$ and $D_{s-1}(a)>0$):
\begin{equation}
  F_s^{(\eta)}(a) \;\coloneqq\; 1 - \eta + \eta\,\frac{W_{\mathrm{act}}(a)\, B_{I_s}(a)}{W_{I_s}\, D_{s-1}(a)},
  \label{eq:multiplier}
\end{equation}
with combined wealth $\mathcal{M}_s^{\mathrm{mix}}(a) \coloneqq \frac{1}{|\mathcal{E}|} \sum_{\eta \in \mathcal{E}} \mathcal{M}_s^{(\eta)}(a)$ (reducing to \texttt{CPPS-Fix} when $|\mathcal{E}|=1$). Rival $a$ is \emph{deterministically refuted} (strict precedence over latching) whenever $D_{s-1}(a) < 0$, $(D_{s-1}(a){=}0 \land B_{I_s}(a)>0)$, or exact intervals $[\underline{T}_c, \overline{T}_c]$ prove $\hat{w}$ beats $a$; otherwise $a$ \emph{statistically latches} if $\mathcal{M}_s^{\mathrm{mix}}(a) \ge 1/\delta$. If $\hat{w}$ is deterministically beaten or $\{W_u > 0\}$ exhausts with unlatched rivals, \emph{chunked survivor completion} queries remaining survivor cells until exact intervals isolate one winner.

\begin{theorem}[Active-Conditioned Exact-Rational \texttt{CPPS-Mix}]
\label{thm:cpps}
Fix integer weights $w_i \ge 1$, tie ranks $r_c$, any $(\hat{S}, S^*) \in \{0,1\}^{N \times M}$, $\delta \in (0,1)$, finite rational grid $\mathcal{E} \subset (0,1)$, and wave cap $b_{\mathrm{cells}} \ge 1$. Under Lazy-Row~\eqref{eq:row_cap} and Atomic-Cell~\eqref{eq:cell_cap} with logical wave ordering and survivor completion:
\begin{enumerate}[label=(\roman*),leftmargin=*,itemsep=1pt,topsep=1pt]
  \item For every rival $a \neq \hat{w}$ satisfying the null $E_a \ge h_a$, each $\mathcal{M}_{s\wedge\tau_a}^{(\eta)}(a)$ and their mixture $\mathcal{M}_{s\wedge\tau_a}^{\mathrm{mix}}(a)$ are nonnegative $\Fcal_s$-supermartingales with $\mathcal{M}_0^{\mathrm{mix}}(a) = 1$.
  \item Inheriting \texttt{SHANGRLA}'s multiplicity-free assertion structure, $\Prob_{\mathrm{audit}}(\tilde{w} \neq w^\star \mid \hat{S}, S^*) \le \delta$ holds with \textbf{zero} $(M{-}1)$ union bound.
\end{enumerate}
\end{theorem}

\subsection{Theorem 2: Causal Envelopes \& Violation-Rate Spot-Check}
\label{sec:thm2}

Two ubiquitous executable proxy constructions induce deterministic one-sided causal envelopes settling $\Kcal_0 \subseteq [N]\times[M]$ at $\Fcal_0$:
\begin{itemize}[leftmargin=1.3em,itemsep=1pt,topsep=1pt]
  \item \textbf{Halting-Conditioned Under-Approximation (\texttt{CPPS-HC}):} Run each candidate under an identical deterministic prefix up to step cap $K < K_{\mathrm{full}}$ (changing \emph{only} \texttt{max\_steps}$=K$). Record halting indicator $H_{ic} \in \{0,1\}$ and gate unhalted scores to $\hat{s}_{ic} = 0$. Prefix invariance yields $H_{ic}=1 \Rightarrow s^*_{ic}=\hat{s}_{ic}$ and $H_{ic}=0 \Rightarrow 0 = \hat{s}_{ic} \le s^*_{ic}$.
  \item \textbf{Necessary-Condition Over-Approximation (\texttt{CPPS-OC}):} Run a necessary test subset (\texttt{base\_input} in \texttt{EvalPlus}); full evaluation runs a superset (\texttt{base\_input} $\cup$ \texttt{plus\_input}), so $s^*_{ic} \le \hat{s}_{ic}$ everywhere ($\hat{s}_{ic}=0 \Rightarrow s^*_{ic}=0$).
\end{itemize}

\begin{theorem}[Causal Envelope Tightening: \texttt{CPPS-HC} / \texttt{CPPS-OC}]
\label{thm:envelope}
Assume $(\hat{S}, H, S^*) \in \mathcal{C}$ is consistent with the active causal envelope, with $\Kcal_0$ pre-seeded into exact score bounds $[\underline{T}_c, \overline{T}_c]$ and fail-closed runtime verification on queried cells (raising \texttt{CausalProxyViolationError} and failing closed to deterministic full completion at risk $0$, or conditional proxy-free \texttt{WOR-CS LUCB} at risk $\le \delta$):
\begin{enumerate}[label=(\alph*),leftmargin=1.6em,itemsep=1pt,topsep=1pt]
  \item \textbf{(\texttt{CPPS-HC})} Replacing~\eqref{eq:row_cap}--\eqref{eq:cell_cap} with $U_i^{\mathrm{HC}}(a) \coloneqq w_i(1{-}H_{ia})$, $U_{i\hat{w}}^{\mathrm{HC}}(a) \coloneqq 0$, $U_{ia}^{\mathrm{HC}}(a) \coloneqq w_i(1{-}H_{ia})$ satisfies $e_u(a) \le U_u^{\mathrm{HC}}(a)$, queries \textbf{zero} halted cells (under Atomic-Cell, or Lazy-Row executing only unhalted cells) and zero $\hat{w}$ cells during Atomic-Cell rival auditing, matches interval dominance identically on the $D < 0$ branch, and satisfies~\eqref{eq:rla} over $\mathcal{C}$.
  \item \textbf{(\texttt{CPPS-OC})} Replacing~\eqref{eq:row_cap}--\eqref{eq:cell_cap} with $U_i^{\mathrm{OC}}(a) \coloneqq w_i \hat{s}_{i\hat{w}}$, $U_{i\hat{w}}^{\mathrm{OC}}(a) \coloneqq w_i \hat{s}_{i\hat{w}}$, $U_{ia}^{\mathrm{OC}}(a) \coloneqq 0$ satisfies $e_u(a) \le U_u^{\mathrm{OC}}(a)$, queries \textbf{zero} $\hat{s}_{ic}=0$ cells (and zero rival cells during Atomic-Cell $\hat{w}$ auditing), matches interval dominance on $D < 0$, and satisfies~\eqref{eq:rla} over $\mathcal{C}$.
\end{enumerate}
\end{theorem}

\paragraph{Engine-agnostic \texttt{WOR-CS+Env} and violation-rate spot-checking.} Runtime guards check only \emph{queried} cells, not skipped $\Kcal_0$ cells. Corollary~\ref{cor:env_spot} both transfers $\Kcal_0$ pre-seeding to confidence-sequence bandits and provides a frequentist hypergeometric test for silent $\Kcal_0$ violations:

\begin{corollary}[\texttt{WOR-CS+Env} \& Spot-Check]
\label{cor:env_spot}
\mbox{}
\begin{enumerate}[label=(\roman*),leftmargin=1.8em,itemsep=1pt,topsep=1pt]
  \item Under Theorem~\ref{thm:envelope}'s admissible class $\mathcal{C}$, pre-seeding $\Kcal_0$ and running level-$\delta/M$ exact-rational WOR betting confidence sequences \citep{waudbysmith2024estimating} (scaled to $[0, w_{\max}]$) over each arm's unsettled pool $\Ucal_c$ with tie-aware LUCB stopping ($\mathrm{LCB}_\ell \ge \mathrm{UCB}_b + \Ind_{r_\ell > r_b}$ for leader $\ell$) certifies~\eqref{eq:rla} at risk $\le \delta$ (\texttt{WOR-CS+Env}).
  \item \textbf{(Hypergeometric Envelope Spot-Check)} Fix an unknown count $V \in \{0,\dots,N_0\}$ ($N_0 = |\Kcal_0|$) of violated cells in $\Kcal_0$, and draw $K_{\mathrm{spot}}$ cells uniformly without replacement, observing $X$ violations: (a) on $\{X \ge 1\}$, discarding $\Kcal_0$ and running conditional proxy-free \texttt{WOR-CS LUCB} at level $\delta_{\mathrm{aud}}$ over unqueried cells guarantees $\Prob(\tilde{w} \neq w^\star \mid \Fcal_{\mathrm{spot}}, X \ge 1) \le \delta_{\mathrm{aud}}$; (b) define $v_{\max}(K_{\mathrm{spot}}, \delta_{\mathrm{spot}})$ as the largest integer $v \ge 0$ with $\binom{N_0-v}{K_{\mathrm{spot}}} / \binom{N_0}{K_{\mathrm{spot}}} > \delta_{\mathrm{spot}}$ ($0$ if $K_{\mathrm{spot}}=N_0$). For every fixed $V > v_{\max}$, $\Prob_V(X = 0) \le \delta_{\mathrm{spot}}$. \emph{Scope:} testing $V \le v_{\max}$ does not alone certify $\Prob(\tilde{w} \neq w^\star) \le \delta$ without either assuming $V=0$ or inflating every arm's bounds by $\pm v_{\max} w_{\max}$ at composed risk $\delta_{\mathrm{aud}} + \delta_{\mathrm{spot}}$ (conservative on tight margins).
\end{enumerate}
\end{corollary}

\subsection{Theorem 3: \texttt{GateAudit-Unified} (Adaptive Re-Anchoring)}
\label{sec:thm3}

Theorem~\ref{thm:cpps} avoids an $M$-way union bound when $\hat{w} = w^\star$, while \texttt{WOR-CS+Env} dynamically switches leaders when $\hat{w} \neq w^\star$ at the cost of a $\delta/M$ threshold. \textbf{\texttt{GateAudit-Unified}} (Algorithm~\ref{alg:unified}) hedges between both regimes via an exact-rational split $\delta = \delta_1 + \delta_2$ (default $\delta_1 = \frac{4}{5}\delta$, $\delta_2 = \frac{1}{5}\delta$):
\begin{enumerate}[leftmargin=1.4em,itemsep=1pt,topsep=1pt]
  \item \textbf{Phase~A (Multiplicity-Free \texttt{CPPS-Mix} at $\delta_1$):} Run Atomic-Cell \texttt{CPPS-Mix} anchored at $\hat{w}$ with threshold $1/\delta_1$. Whenever a rival $a$'s mixture wealth drops to $\mathcal{M}^{\mathrm{mix}}_s(a) \le \theta_{\mathrm{sus}}$ ($\frac{1}{4}$) after $\ge n_{\min}$ ($4$) active draws, immediately pause $a$ into $\mathcal{P}$ and remove $a$ from $\Rcal$ (freezing its wealth so inverted proxies stop spending waves on losing bets) while continuing to refute clearly dominated rivals.
  \item \textbf{Phase~B (Conditional Survivor Re-Anchoring at $\delta_2$):} At Phase~A stopping time $\tau$ with survivors $\Scal_\tau \subseteq [M]$, if $\Scal_\tau = \{\hat{w}\}$, return $\hat{w}$. Else, conditioned on $\Fcal_\tau$, draw a \textbf{fresh independent WOR permutation} over each survivor's remaining unknown pool $\Ucal_c^{(\tau)}$ and run exact-rational \texttt{WOR-CS LUCB} at level $\delta_2 / |\Scal_\tau|$.
\end{enumerate}

\begin{theorem}[\texttt{GateAudit-Unified} Exact Finite-Sample Validity]
\label{thm:unified}
Fix weights $w_i \ge 1$, tie ranks $r_c$, any $(\hat{S}, H, S^*) \in \mathcal{C}$, rational split $\delta_1 + \delta_2 = \delta$ ($\delta_1, \delta_2 > 0$), finite grid $\mathcal{E} \subset (0,1)$, and $\Fcal_{s}$-measurable suspect rule. Then Algorithm~\ref{alg:unified} satisfies $\Prob_{\mathrm{audit}}(\tilde{w} \neq w^\star \mid \hat{S}, H, S^*) \le \delta_1 + \delta_2 = \delta$ in exact \texttt{Fraction} arithmetic.
\end{theorem}

\emph{Proof sketch (Appendix~\ref{app:proofs_main}).} Let $\mathcal{E}_1 \coloneqq \{w^\star \notin \Scal_\tau\}$. Pausing never eliminates an arm, so by Theorem~\ref{thm:cpps} on the single null $\mathcal{M}^{\mathrm{mix}}_s(w^\star)$, $\Prob(\mathcal{E}_1) \le \delta_1$. On $\mathcal{E}_1^c$, Phase~B conditions on $\Fcal_\tau$ and draws a fresh independent WOR permutation over $\{\Ucal_c^{(\tau)}\}_{c \in \Scal_\tau}$, yielding $\Prob(\tilde{w} \neq w^\star \mid \Fcal_\tau) \le \delta_2$ by Corollary~\ref{cor:env_spot}(i). Thus $\Prob(\tilde{w} \neq w^\star) \le \delta_1 + \delta_2 = \delta$. \qed

\begin{algorithm}[t]
\caption{\texttt{GateAudit-Unified} (Theorems~\ref{thm:cpps}--\ref{thm:unified})}
\label{alg:unified}
\scriptsize
\begin{algorithmic}[1]
  \STATE {\bfseries Input:} Proxy $\hat{S}$, mask $H$, mode, weights $w$, ranks $r$, risk $\delta = \delta_1{+}\delta_2$, grid $\mathcal{E}$, wave cap $b$, suspect params $(\theta_{\mathrm{sus}}, n_{\min})$.
  \STATE Pre-seed $\Kcal_0$ in exact bounds $[\underline{T}_c, \overline{T}_c]$; $\hat{w} \gets \arg\min_c(-\hat{T}_c, r_c)$; active $\Rcal \gets [M]\setminus\{\hat{w}\}$; suspects $\mathcal{P} \gets \emptyset$; exact \texttt{Fraction} $\mathcal{M}^{(\eta)}(a) \gets 1$.
  \WHILE{$\Rcal \neq \emptyset$ {\bfseries and} $\{u \in P : W_u > 0\} \neq \emptyset$}
    \STATE {\bfseries if} $[\underline{T}_c, \overline{T}_c]$ isolates exact global winner $g^\star$ {\bfseries then return} $g^\star$
    \STATE Draw ordered wave $(I_{t+1},\dots,I_{t+k})$ ($k \le b$) via exact integer-ticket PPS-WOR on $\{W_u > 0\}$; query oracle in parallel and verify envelope (fail-closed fallback on violation).
    \FOR{$s = t+1$ {\bfseries to} $t+k$ (pre-committed logical order)}
      \STATE Update $[\underline{T}_c, \overline{T}_c]$; for each $a \in \Rcal$ with $I_s \in A_{s-1}(a)$:
      \STATE \quad {\bfseries if} $D_{s-1}(a) < 0$ {\bfseries or} $(D_{s-1}(a){=}0 \land B_{I_s}(a){>}0)$ {\bfseries then} refute $a$;
      \STATE \quad {\bfseries else} update $\mathcal{M}^{(\eta)}(a)$ via~\eqref{eq:multiplier}; $\mathcal{M}^{\mathrm{mix}}(a) \gets \frac{1}{|\mathcal{E}|}\sum_\eta \mathcal{M}^{(\eta)}(a)$;
      \STATE \quad \quad latch $a$ if $\mathcal{M}^{\mathrm{mix}}(a) \ge 1/\delta_1$; else if $\mathcal{M}^{\mathrm{mix}}(a) \le \theta_{\mathrm{sus}}$ after $\ge n_{\min}$ draws, move $a$ immediately from $\Rcal$ to $\mathcal{P}$.
    \ENDFOR
    \STATE Prune refuted/latched arms from $\Rcal$.
  \ENDWHILE
  \STATE $\Scal_\tau \gets$ all non-eliminated arms (incl.\ $\hat{w}$ if not beaten, $\mathcal{P}$, unlatched).
  \STATE {\bfseries if} $|\Scal_\tau| = 1$ {\bfseries then return} sole member of $\Scal_\tau$.
  \STATE \textbf{Phase B:} Draw fresh independent WOR permutation over remaining unknown pools $\{\Ucal_c^{(\tau)}\}_{c \in \Scal_\tau}$; run exact-rational \texttt{WOR-CS LUCB} at level $\delta_2 / |\Scal_\tau|$ until exact bounds or CS separation certifies $\tilde{w}$.
\end{algorithmic}
\end{algorithm}

\paragraph{Exact-rational numerical safety and latency.} Naive IEEE-754 \texttt{float64} arithmetic (without outward-rounded interval or directed-rounding log arithmetic) risks rounding errors at extreme weights or $D{=}0$ boundaries. GateAudit runs all wealth updates, draws, and bounds in Python \texttt{fractions.Fraction} ($2{,}000$ randomized adversarial trials, $18$ exact draw-tree enumerations over $892$ paths, and $97/97$ unit tests with $0$ risk violations; Appendix~\ref{app:provenance}), taking only $124.1\text{ ms}$ mean controller wall-time ($405.5\text{ ms}$ for \texttt{CPPS-Mix}; peak precision $\le 2{,}676$ bits) on $20{,}000$-cell \texttt{SWE-bench} $M{=}40$.

\section{Systems Architecture: Forking, Continuation, and Stage-0 Routing}
\label{sec:systems}

\TableOneLiveGPU

\begin{figure*}[t]
  \centering
  \includegraphics[width=0.80\textwidth]{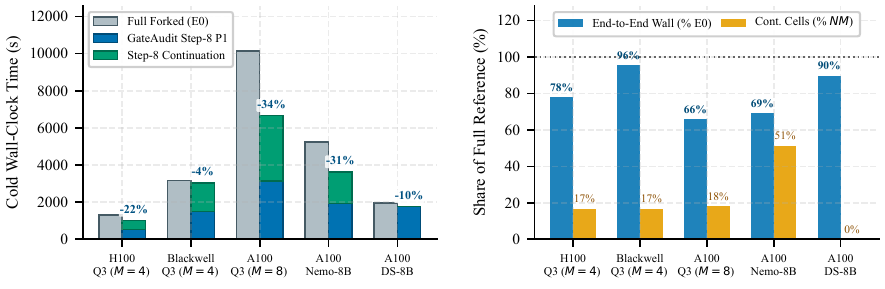}
  \vspace{-0.9em}
  \caption{\textbf{Live closed-loop \texttt{AppWorld} \texttt{dev} wall-clock evaluation ($n{=}1$/config, fixed E0$\to$E2 order).} \textbf{(a)} Cold wall-clock breakdown (Full Forked E0 vs.\ Step-8 P1 + Continuation E2, verified \texttt{/flush\_cache}). \textbf{(b)} End-to-end wall-clock (\% E0) vs.\ continuation cells (\% $NM$); \texttt{DS-8B} ($72/72$ halted, $0/72$ mismatches) acts as an A/A check measuring a $10.1\%$ first-run cold-order bias.}
  \label{fig:live_gpu}
  \vspace{-0.8em}
\end{figure*}

\paragraph{Isolated workers, Turn-1 forking, and Step-$K$ continuation.} Mutable worlds run in persistent \texttt{spawn} workers verified by SQLite digests (\texttt{appworld\_db\_state\_v3}); cold GPU boundaries verify \texttt{SGLang} \texttt{RadixCache} eviction via sentinel probes (\texttt{/flush\_cache}). Shared-Turn-1 harnesses run Turn~1 once per task and replay verified actions into siblings ($-20.1\%$ cold \texttt{H100} makespan). Under \texttt{CPPS-HC}, audited unhalted cells replay token-hash-verified steps $1\dots K$ locally into fresh SQLite workers and decode live from step $K{+}1$ over warm \texttt{RadixCache}; without continuation, re-running from scratch on \texttt{H100} calibration was $+13.5\%$ \emph{slower} than full evaluation.

\paragraph{Pre-audit gate and Stage-0 routing.} On a disjoint calibration split, \texttt{ProxyFallbackTracker} rejects $\hat{S}$ on column collapse ($>85\%$ identical rows), $>50\%$ pairwise disagreement, or tied leaders. Otherwise, \texttt{ProspectiveSelectorModel} fits an envelope-respecting Dirichlet transition law and prices Stage-0 (\emph{proxy vs.\ direct forked full}) and Stage-1 (\emph{Lazy-Row vs.\ Atomic-Cell vs.\ unknown-only}) via Monte Carlo simulation on measured $p$, continuation multipliers, and API/compute costs---affecting cost only, never $\delta$-risk.

\section{Empirical Evaluation}
\label{sec:experiments}

\TableTwoBenchmarks

\begin{figure*}[t]
  \centering
  \includegraphics[width=0.80\textwidth]{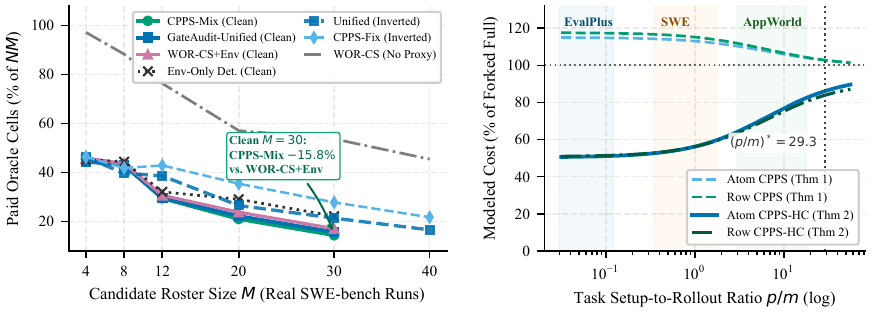}
  \vspace{-0.9em}
  \caption{\textbf{Real $M{=}40$ \texttt{SWE-bench Verified} scaling and modeled unit crossover.} \textbf{(a)} Paid oracle cells (\% $NM$) across $M \in \{4,\dots,40\}$ real \texttt{mini-swe-agent} runs ($N{=}500$, $K{=}40$ counterfactual truncation, $\delta{=}0.05$, $0.0\%$ err). Across all $16$ $M$-configs (\texttt{Frontier}, \texttt{Rand-10}, curated \texttt{Clean}), \texttt{Unified} caps worst-case empirical regret vs.\ $\min(\texttt{CPPS-Mix}, \texttt{WOR-CS+Env})$ at $8.8\%$ (vs.\ $18.8\%$ \texttt{WOR-CS+Env}, $31.1\%$ \texttt{CPPS-Mix}). \textbf{(b)} Modeled unit cost vs.\ $p/m$ illustrating Remark~\ref{rem:reversal} ($(p/m)^\star = 29.3$, $M{=}8$).}
  \label{fig:m40_phase}
  \vspace{-0.8em}
\end{figure*}

\begin{figure*}[t]
  \centering
  \includegraphics[width=0.75\textwidth]{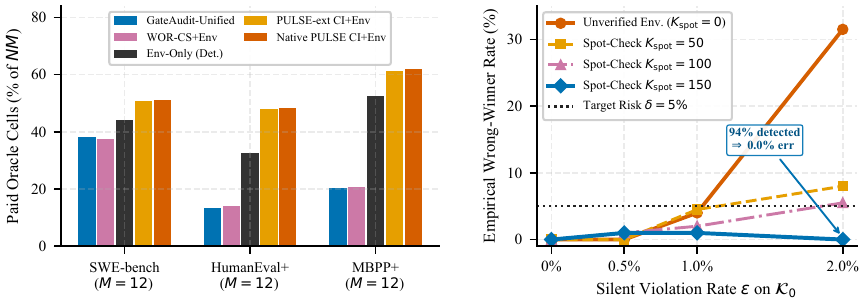}
  \vspace{-1.0em}
  \caption{\textbf{Apples-to-apples \texttt{+Env} comparison and Hypergeometric Spot-Check.} \textbf{(a)} Mean paid cells (\% $NM$) at $M{=}12$ under identical $\Kcal_0$: \texttt{Unified} uses $1.33{\times}\text{--}3.60{\times}$ fewer cells than \texttt{Native PULSE CI-stop+Env}. \textbf{(b)} Under silent violation rate $\epsilon \in \{0,\dots,2\%\}$ on skipped $\Kcal_0$ cells (\texttt{SWE-bench} $M{=}4$, $|\Kcal_0|{=}1{,}043$, $\delta_{\mathrm{spot}}{=}0.01$, $200$ seeds), unverified error hits $31.5\%$, while $K_{\mathrm{spot}}{=}150$ detects $94.0\%$ and cuts error to $0.0\%$ ($K_{\mathrm{spot}}{\in}\{50,100\}$ exceed $5\%$ at $\epsilon{=}2\%$; Table~\ref{tab:app_spotcheck}).}
  \label{fig:frontier_spot}
  \vspace{-1.0em}
\end{figure*}

We evaluate GateAudit on: (1) five \textbf{live closed-loop GPU runs} on \texttt{AppWorld dev} \citep{trivedi2024appworld} ($6$ disjoint eval families $= 18$ tasks; \texttt{test\_normal}/\texttt{test\_challenge} sealed) across \texttt{H100} (\texttt{sm\_90}), \texttt{Blackwell} (\texttt{sm\_120}), and \texttt{A100} (\texttt{sm\_80}) with \texttt{Qwen3-8B} \citep{yang2025qwen3} and \texttt{Llama-3.1} \citep{grattafiori2024llama3,deepseek2025r1,bercovich2025nemotron} under deterministic \texttt{SGLang v0.5.10.post1}; (2) \textbf{$M{=}40$ \texttt{SWE-bench Verified}} (counterfactual $K{=}40$ truncation, $N{=}500$) and locally timed \textbf{\texttt{EvalPlus}} (\texttt{HumanEval+} $N{=}164$, \texttt{MBPP+} $N{=}378$) with \texttt{+Env} baselines; and (3) \textbf{stress tests} (spot-checking, near-null $\delta$, sensitivity, USD vs.\ ms, synthetic stochastic rollouts).

\subsection{Live Multi-GPU \& Multi-Architecture \texttt{AppWorld} Study}
\label{sec:exp_live}

On \texttt{H100}, Turn-1 forking ($24$ cells/arm, $C_0\text{--}C_3$) cuts cold makespan $20.1\%$ ($495.9\text{s}\!\to\!396.2\text{s}$, $24/24$ exact match). On the $6$-family calibration split ($R_0\text{--}R_3$, Full $[4,3,2,2]$), disabling \texttt{<think>} caused $11/13$ ($84.6\%$) task-level candidate-pair disagreements (rejected by \texttt{ProxyFallbackTracker}), while Step-8 halting (\texttt{Path E}, $-52.2\%$ P1 time) settled $45/72$ cells ($0$ mismatches).

Table~\ref{tab:live_gpu} and Figure~\ref{fig:live_gpu} report five $n{=}1$ live disjoint evaluations (fixed \texttt{E0}$\to$\texttt{E2} order). Critically, \emph{none} of the live A100 runs triggered statistical latching (\texttt{stat\_elim=False}): live GPU savings validate \textbf{Theorem~\ref{thm:envelope}'s causal envelope and Step-8 continuation}, while Theorems~\ref{thm:cpps} and~\ref{thm:unified}'s statistical layers are exercised at larger $N, M$ in \S\ref{sec:exp_external}:
\begin{itemize}[leftmargin=1.3em,itemsep=0.5pt,topsep=1pt]
  \item \textbf{A/A order-bias checks \& order-adjusted savings:} Since \texttt{E0} always preceded \texttt{E2}, two A/A checks bound first-run bias (cold start / Triton autotuning): on \texttt{H100}, Stage-0 \texttt{E1} routed to full forked evaluation ($1{,}264.3\text{s}$ vs.\ \texttt{E0} $1{,}302.5\text{s}$, $2.9\%$ gap), giving a bias-robust \textbf{$-19.8\%$} \texttt{H100} saving (\texttt{E2} $1{,}013.5\text{s}$, $-22.2\%$ vs.\ \texttt{E0}; $12/14$ unhalted; $42.1\%$ warm \texttt{RadixCache}). On \texttt{A100} \texttt{DS-8B}, all $72/72$ cells halt before step~8 ($0/72$ mismatches), revealing a $10.1\%$ first-run cold-order bias ($1{,}960.3\text{s}\!\to\!1{,}761.7\text{s}$). Subtracting $10.1\%$ leaves \texttt{Blackwell}'s raw $4.5\%$ gap ($3{,}180.7\text{s}\!\to\!3{,}038.1\text{s}$; $12/14$ unhalted) net negative, while preserving \textbf{$\approx 21\%\text{--}24\%$ order-adjusted \texttt{A100} savings}: \texttt{Nemotron-Nano-8B} (raw $-30.8\%$, $5{,}245.1\text{s}\!\to\!3{,}629.5\text{s}$, $0/24$ halted mismatches, $37/37$ continuations; offline 200-seed \texttt{Unified} $36.74$ vs.\ \texttt{WOR-CS+Env} $48.0$, \texttt{CPPS-Fix/Mix} $36.0$) and \texttt{Qwen3-8B} $M{=}8$ (raw $-34.2\%$, $10{,}130\text{s}\!\to\!6{,}665\text{s}$, $26/144$ continuations; offline \texttt{Unified} $26.64$ vs.\ \texttt{WOR-CS+Env} $29.30$, \texttt{CPPS-Fix} $26.12$).
  \item \textbf{Within-stack drift vs.\ cross-hardware $S^*$ shift:} Only \texttt{H100} has a strict winner ($[6,5,7,6]$, $R_2$); both $8\text{B}$ \texttt{Llama-3.1} runs score $0/18$ (zero-signal tie-break tests). Separate from \emph{within-stack batch drift}---$0/58$ halted and $12/12$ continuation matches on \texttt{H100} and \texttt{Blackwell} \texttt{E0}$\to$\texttt{E2} ($1/63{=}1.6\%$ on \texttt{Blackwell} \texttt{Row Poll}), vs.\ $1/107$ ($0.9\%$) halted and $3/26$ ($11.5\%$) continuation drift on high-batch \texttt{A100} $M{=}8$ ($[7,8,6,2,8,8,7,7]$, $3$-way tie)---\texttt{Blackwell} vs.\ \texttt{H100} on the same $18$ tasks showed \emph{cross-hardware FP divergence} flipping $5/18$ $R_0$ trajectories into a $[7,7,4,6]$ tie ($w^\star{=}R_0$).
  \item \textbf{Full $57$-task \texttt{dev} trace ($228$ cells, $C_0\text{--}C_3$):} At a $1$-task margin ($[15,15,15,16]$), \texttt{CPPS-HC} queries all $47/228$ unhalted cells ($228{-}181$, $-79.4\%$, $13/57$ tasks, $-33.6\%$ wave time, $0\%$ err; pure causal-envelope savings).
\end{itemize}

\subsection{Apples-to-Apples Multi-Benchmark \& $M{=}40$ Scaling}
\label{sec:exp_external}

\paragraph{Apples-to-apples \texttt{+Env} baselines (Table~\ref{tab:benchmarks}, Figure~\ref{fig:frontier_spot}a).} Every \texttt{+Env} comparator---\texttt{Env-Only} (deterministic sequential evaluation of unsettled cells), \texttt{WOR-CS+Env} (Corollary~\ref{cor:env_spot}(i)), \texttt{Native PULSE CI-stop+Env} (Welford variance, Jeffreys add-$\tfrac{1}{2}$ floor, no double FPC, Bonferroni CI stopping), \texttt{PULSE-ext CI-stop+Env} (Lemma~\ref{lem:pulse_ext}), \texttt{SH+Env} \citep{karnin2013halving}, and \texttt{UCB-E+Env} \citep{audibert2010best}---receives the \emph{exact same} $\Kcal_0$. (\texttt{SWE-bench} $K{=}40$ truncates logged trajectories, so $0.0\%$ violation holds by construction.) At $M{=}12$, \texttt{Unified} cuts cells $1.33{\times}\text{--}3.60{\times}$ vs.\ \texttt{Native PULSE CI-stop+Env} ($38.3\%/13.4\%/20.4\%$ vs.\ $51.0\%/48.3\%/62.1\%$, and $2.32{\times}\text{--}6.20{\times}$ vs.\ no-env \texttt{PULSE}; at $M{=}4$, $0.99{\times}\text{--}1.66{\times}$, where \texttt{PULSE+Env} uses $45.5\%$ at $1.0\%$ err vs.\ \texttt{Unified} $46.1\%$ at $0.0\%$ err on \texttt{SWE-bench}). On locally timed \texttt{EvalPlus}, \texttt{Unified} cuts net runtime $73.0\%$ (\texttt{HE+}) and $72.9\%$ (\texttt{MBPP+}) at $M{=}12$ ($47.6\%/53.6\%$ at $M{=}4$), beating \texttt{WOR-CS+Env} at $M{=}12$ (paired 95\% CIs: $[-14.6, -7.7]$ on \texttt{HE+}; $[-22.4, -6.2]$ on \texttt{MBPP+}; tying on \texttt{HE+} $M{=}4$, $+0.21$, CI $[-3.78, +4.21]$), while fixed-budget \texttt{SH+Env}/\texttt{UCB-E+Env} err up to $54.5\%/47.5\%$ (note \texttt{SH+Env} under-spends on \texttt{SWE-bench}: $634/926$ at $M{=}4$, $1{,}282/2{,}321$ at $M{=}12$).

\paragraph{Real $M{=}40$ \texttt{SWE-bench Verified} scaling \& empirical worst-case regret (Figure~\ref{fig:m40_phase}a, Table~\ref{tab:app_m40}).} Across $40$ official \texttt{mini-swe-agent} runs ($20{,}000$ cells, $K{=}40$, $0.0\%$ err):
\begin{itemize}[leftmargin=1.3em,itemsep=0.5pt,topsep=1pt]
  \item \textbf{Unbiased rosters (\texttt{Rand-10} \& \texttt{Frontier}):} Over $10$ uniform random $M$-subsets ($\times 10$ seeds each; inverted $7/10\text{--}10/10$ times), \texttt{Unified} saves $30.3\%$ vs.\ \texttt{Env-Only} at $M{=}30$ ($2{,}582.2$ vs.\ $3{,}707.1$). On proxy-inverted \texttt{Frontier} ($M{=}40$), \texttt{Unified} uses $\SweFrontMFortyUniCells$ cells ($-83.4\%$ vs.\ full, $-1{,}031.2$ vs.\ \texttt{CPPS-Fix} $\SweFrontMFortyFixCells$, $2.74{\times}$ vs.\ \texttt{WOR-CS} $\SweFrontMFortyWorcsCells$). On all $11$ \texttt{Rand-10}/\texttt{Frontier} configs, our \texttt{WOR-CS+Env} edges out \texttt{Unified} by $0.3\%\text{--}4.2\%$ (and significantly at $M{\in}\{12,20,30\}$ up to $6.4\%$ at $M{=}20$, CI $[44.9, 312.7]$, on a $40$-seed near-frontier slice); \texttt{CPPS-Mix} improves sharply over \texttt{CPPS-Fix} at $M \ge 20$ ($-15.3\%$ at $M{=}40$) while slightly trailing at $M \in \{4, 12\}$ ($936$ vs.\ $928$; $2{,}591$ vs.\ $2{,}574$).
  \item \textbf{Curated \texttt{Clean} roster \& worst-case regret hedge:} To isolate $\hat{w} = w^\star$, \texttt{Clean} pairs the highest-scoring Step-$40$-consistent model ($\hat{w}{=}w^\star$) with all strictly weaker models (using $S^*$). At $M{=}30$, \texttt{CPPS-Mix} uses $\SweCleanMThirtyMixCells$ cells ($-15.8\%$ vs.\ \texttt{WOR-CS+Env} $\SweCleanMThirtyWceCells$, $-35.1\%$ vs.\ \texttt{Env-Only}), and \texttt{Unified} uses $\SweCleanMThirtyUniCells$ ($-8.4\%$ vs.\ \texttt{WOR-CS+Env}). Across all $16$ \texttt{SWE-bench} $M$-configs, \textbf{\texttt{Unified}'s worst-case empirical regret vs.\ $\min(\texttt{CPPS-Mix}, \texttt{WOR-CS+Env})$ is $8.8\%$}, vs.\ \textbf{$18.8\%$ for \texttt{WOR-CS+Env}} and \textbf{$31.1\%$ for \texttt{CPPS-Mix}}---an empirical cross-regime hedge, not an unconditional winner.
\end{itemize}

\paragraph{Honest USD vs.\ cell economics (Table~\ref{tab:app_kusd}).} Separating API USD (\texttt{SWE-bench}) from ms (\texttt{EvalPlus}) reveals a clear boundary: when fresh Step-$K$ tokens ($K{\in}\{10,15,20,40\}$) are billed upfront without sunk prefixes, proxy-free \texttt{WOR-CS LUCB} spends fewer USD at $M{=}12$ ($\$2{,}046$ vs.\ $\$2{,}149\text{--}\$2{,}289$) and $M{=}40$ ($\$4{,}060$ vs.\ $\$4{,}905\text{--}\$6{,}274$), motivating Stage-0 gating. At $M{=}4, K{=}10$ (\texttt{Unified} $\$846$ vs.\ $\$853$) or with sunk/cached trajectories, GateAudit saves $28.4\%\text{--}89.1\%$ of tail USD ($72.9\%\text{--}73.0\%$ \texttt{EvalPlus} ms at $M{=}12$, $47.6\%\text{--}53.6\%$ at $M{=}4$).

\subsection{Stress Tests: Spot-Check, Calibration, \& Rollouts}
\label{sec:exp_stress}

\begin{itemize}[leftmargin=1.3em,itemsep=0.5pt,topsep=1pt]
  \item \textbf{Hypergeometric Spot-Check (Figure~\ref{fig:frontier_spot}b, Table~\ref{tab:app_spotcheck}):} Injecting $\epsilon{\in}\{0\%, 0.5\%, 1\%, 2\%\}$ violations on skipped $\Kcal_0$ cells (\texttt{SWE-bench} $M{=}4$, $|\Kcal_0|{=}1{,}043$, $\delta_{\mathrm{spot}}{=}0.01$, $\delta_{\mathrm{aud}}{=}0.05$, $200$ seeds) drives unverified error ($K_{\mathrm{spot}}{=}0$) to $31.5\%$ at $\epsilon{=}2\%$; $K_{\mathrm{spot}}{\in}\{50,100\}$ exceed $5\%$ ($8.0\%, 5.5\%$), while $K_{\mathrm{spot}}{=}150$ rejects $V{\ge}30$ ($2.88\%$), detects $94.0\%$, and cuts error to \textbf{$0.0\%$}.
  \item \textbf{Near-null calibration ($500$ seeds, $\delta{=}5\%$):} On a $1$-cell-margin near-null ($N{=}120, M{=}8$), error rates are consistent with $\delta{=}5\%$: $\NearNullFixErr$ (\texttt{CPPS-Fix}, 95\% CP upper $6.45\%$), $\NearNullMixErr$ (\texttt{CPPS-Mix}, $4.82\%$), $\NearNullUniErr$ (\texttt{Unified}, $3.12\%$), $0.20\%$ (\texttt{WOR-CS+Env}, $0.95\%$).
  \item \textbf{Hyperparameter sensitivity ($36$ configs, $25$ seeds/config):} Sweeping $\delta_1/\delta{\in}\{\frac{1}{2},\frac{2}{3},\frac{4}{5},\frac{9}{10}\}$, $\theta_{\mathrm{sus}}{\in}\{\frac{1}{8},\frac{1}{4},\frac{1}{2}\}$, $n_{\min}{\in}\{2,4,8\}$ around default $(\frac{4}{5},\frac{1}{4},4)$ shifts \texttt{Unified} $-6.1\%/+3.8\%$ on \texttt{SWE} $M{=}12$ ($\SensSweMin\text{--}\SensSweMax$, $0/25$ err) and $-0.4\%/+1.7\%$ on \texttt{MBPP+} $M{=}12$ ($\SensMbppMin\text{--}\SensMbppMax$, $\le 4\%$ err).
  \item \textbf{Synthetic stochastic rollouts (Table~\ref{tab:app_stochastic}):} Over $300$ trials ($N{=}120, M{=}4, \delta_{\mathrm{audit}}{=}0.025$, $R{\in}\{1,3,5\}$), GateAudit achieves $0.00\%$ matrix error (matching population error $3.33\%/0.33\%/0.00\%$), saving $76.3\%\text{--}\StochRFiveSav$ of cells.
\end{itemize}

\section{Related Work, Limitations, and Conclusion}
\label{sec:conclusion}

\paragraph{Related work.} Coresets \citep{polo2024tinybenchmarks,vivek2024anchor,perlitz2024efficient} lack certificates; PPI \citep{angelopoulos2023ppi,boyeau2024autoeval,chatzi2024ppr,zrnic2024active} gives asymptotic CIs; \texttt{PULSE} \citep{pulse2026active} gives fixed-budget CIs without anytime stopping; bandits \citep{kalyanakrishnan2012lucb,karnin2013halving,audibert2010best,garivier2016track} and judge routing \citep{ao2026service} omit causal envelopes. GateAudit builds on election RLAs (\texttt{SHANGRLA} \citep{stark2020shangrla}, PPEB \citep{aslam2008ppeb}, \citep{stark2010supersimple}, \texttt{RiLACS} \citep{waudbysmith2021rilacs}, \texttt{SUITE} \citep{spertus2022suite}, \texttt{ALPHA} \citep{stark2023alpha}) and betting CSs \citep{waudbysmith2020wor,waudbysmith2024estimating,howard2021time,ramdas2023game}, adding causal envelopes, adaptive re-anchoring, and continuation \citep{zheng2024sglang,he2025defeating,pan2026agentreplay,xin2026detent}.

\paragraph{Limitations \& conclusion.} (1) Realized-matrix certificates bind to one hardware/kernel stack: cross-GPU FP differences flipped the \texttt{AppWorld} winner (\texttt{H100} $R_2$ vs.\ \texttt{Blackwell} $R_0$), while \emph{within-stack} Triton drift ($1/107$ halted, $3/26$ continuation on \texttt{sm\_80} $M{=}8$; $1/63$ on \texttt{sm\_120} \texttt{Row Poll}) requires batch-invariant kernels (two-sided Theorem~\ref{thm:cpps} handles prefix drift only, not oracle drift). (2) Live \texttt{AppWorld} runs are $n{=}1$/config in fixed \texttt{E0}$\to$\texttt{E2} order ($10.1\%$ \texttt{A100} / $2.9\%$ \texttt{H100} \texttt{E1} A/A bias), tie-dominated ($4/5$), with zero statistical latching (interleaved repeats and live $\mathrm{temp}>0$ remain future work); \texttt{SWE-bench} $K$-halting is counterfactual replay. (3) Unsunk frontier API prefixes at large $M$ favor proxy-free \texttt{WOR-CS LUCB} (Table~\ref{tab:app_kusd}). Overall, GateAudit's causal envelopes, re-anchoring ($8.8\%$ max regret), and continuation cut cells $54\%\text{--}87\%$ ($K{=}40$/\texttt{EvalPlus}) and GPU wall-clock $19.8\%$ vs.\ \texttt{H100} \texttt{E1} ($\approx 21\%\text{--}24\%$ order-adjusted on \texttt{A100}).\label{end_of_main}

\section*{Impact Statement}
This work reduces the compute, API spend, and energy footprint of iterative language-agent and code-model evaluation while maintaining rigorous finite-sample statistical error bounds on well-specified deterministic matrices. Potential risks include misapplying causal envelopes outside valid prefix-invariant or subset-test regimes in safety-critical evaluations---mitigated (though not eliminated without bound inflation) by our fail-closed queried-cell runtime guard (\texttt{CausalProxyViolationError}), \texttt{ProxyFallbackTracker}, and frequentist Hypergeometric Envelope Spot-Check (Corollary~\ref{cor:env_spot})---or misinterpreting a fixed-matrix certificate as cross-hardware or unseen-task generalization. All experiments respect benchmark maintainer policies, including holding out protected \texttt{AppWorld} test splits.


\newpage
\appendix
\onecolumn
\renewcommand{\thetable}{A\arabic{table}}
\setcounter{table}{0}

\section{Complete Proofs of Theorems~\ref{thm:cpps}--\ref{thm:unified} and Corollary~\ref{cor:env_spot}}
\label{app:proofs_main}

\subsection{Proof of Theorem~\ref{thm:cpps} (Active-Conditioned Exact-Rational \texttt{CPPS-Mix})}
\begin{proof}
Fix integer task weights $w_1,\dots,w_N \in \mathbb{Z}_{\ge 1}$, strict permutation tie-break ranks $(r_0,\dots,r_{M-1})$, deterministic matrices $(\hat{S}, S^*) \in \{0,1\}^{N \times M}$, target risk $\delta \in (0, 1)$, finite rational grid $\mathcal{E} \subset (0, 1)$, and wave cap $b_{\mathrm{cells}} \ge 1$. Let $\Ucal = [N]$ (Lazy-Row) or $\Ucal = [N] \times [M]$ (Atomic-Cell).

\paragraph{Step 1: Exactness of tie-aware discrepancy characterization.}
By~\eqref{eq:winner}, rival $a \neq \hat{w}$ strictly precedes $\hat{w}$ iff $T^*_a > T^*_{\hat{w}}$ or ($T^*_a = T^*_{\hat{w}}$ and $r_a < r_{\hat{w}}$). By integrality, $E_a = \sum_{u \in \Ucal} e_u(a) = (\hat{T}_{\hat{w}} - \hat{T}_a) - (T^*_{\hat{w}} - T^*_a) = m_a + (T^*_a - T^*_{\hat{w}}) \in \mathbb{Z}$. Hence $a$ outranks $\hat{w}$ iff $E_a \ge m_a + \Ind_{r_a > r_{\hat{w}}} = h_a$.

\paragraph{Step 2: Pointwise upper-cap validity.}
For Lazy-Row ($u=i$), $-(s^*_{i\hat{w}} - s^*_{ia}) \le 1 \implies e_i(a) \le w_i(1 + \hat{s}_{i\hat{w}} - \hat{s}_{ia}) = U_i(a) \ge 0$, and when $\hat{s}_{i\hat{w}}=0, \hat{s}_{ia}=1$, $e_i(a) = w_i(-1 - s^*_{i\hat{w}} + s^*_{ia}) \le 0 = U_i(a)$. For Atomic-Cell ($u=(i,c)$): if $c=\hat{w}$, $e_{i\hat{w}}(a) = w_i(\hat{s}_{i\hat{w}} - s^*_{i\hat{w}}) \le w_i\hat{s}_{i\hat{w}} = U_{i\hat{w}}(a)$; if $c=a$, $e_{ia}(a) = w_i(s^*_{ia} - \hat{s}_{ia}) \le w_i(1 - \hat{s}_{ia}) = U_{ia}(a)$; else $e_{ic}(a) = 0 = U_{ic}(a)$. Hence $B_u(a) = U_u(a) - e_u(a) \ge 0$ everywhere and $U_u(a)=0 \implies e_u(a) \le 0$.

\paragraph{Step 3: Conditional active law under logical wave ordering.}
Let $\mathcal{O}_{s-1}$ include all previously queried units together with any $\Fcal_0$-settled units $\Kcal_0$ (under Theorem~\ref{thm:envelope}). At wave start, $P$, $\Rcal$, and $W_u = \max_{b \in \Rcal} U_u(b)$ are $\Fcal_t$-measurable. Sequential without-replacement sampling gives $\Prob(I_s = u \mid \Fcal_{s-1}) = W_u / \sum_{v \in P^+_{s-1}} W_v$ for $u \in P^+_{s-1}$. For any active $a \in \Rcal$, $A_{s-1}(a) \subseteq P^+_{s-1}$, so whenever $W_{\mathrm{act}}(a) > 0$,
\[
  \Prob\bigl(I_s = u \;\big|\; \Fcal_{s-1},\, I_s \in A_{s-1}(a)\bigr) \;=\; \frac{W_u}{W_{\mathrm{act}}(a)}, \qquad \forall u \in A_{s-1}(a).
\]
Because updates strictly follow pre-committed logical draw order $(I_{t+1},\dots,I_{t+k})$, this law is invariant to asynchronous worker completion times.

\paragraph{Step 4: Component and mixture supermartingales under $E_a \ge h_a$.}
Assume $E_a \ge h_a$. Partition $\Ucal = \mathcal{O}_{s-1} \cup A_{s-1}(a) \cup Z_{s-1}(a)$ with $U_u(a)=0 \implies e_u(a) \le 0$ on $Z_{s-1}(a)$. Then $\sum_{u \in A_{s-1}(a)} e_u(a) \ge E_a - \sum_{u \in \mathcal{O}_{s-1}} e_u(a) \ge r_{s-1}(a)$, so
\begin{equation}
  D_{s-1}(a) \;=\; U_{\mathrm{rem}}(a) - r_{s-1}(a) \;\ge\; \sum_{u \in A_{s-1}(a)} B_u(a) \;\ge\; 0.
  \label{eq:app_null_slack}
\end{equation}
Therefore: (1) $D_{s-1}(a) < 0$ is impossible under $E_a \ge h_a$; (2) if $D_{s-1}(a) = 0$, $B_u(a) = 0$ for every $u \in A_{s-1}(a)$, so $F_s^{(\eta)}(a) = 1$; (3) if $D_{s-1}(a) > 0$, for every $\eta \in \mathcal{E}$ we have $F_s^{(\eta)}(a) \ge 1 - \eta > 0$ and
\[
  \E\bigl[F_s^{(\eta)}(a) \;\big|\; \Fcal_{s-1},\, I_s \in A_{s-1}(a)\bigr]
  = 1 - \eta + \eta \sum_{u \in A_{s-1}(a)} \frac{W_u}{W_{\mathrm{act}}(a)} \cdot \frac{W_{\mathrm{act}}(a)\, B_u(a)}{W_u\, D_{s-1}(a)}
  = 1 - \eta + \eta\,\frac{\sum_{u \in A_{s-1}(a)} B_u(a)}{D_{s-1}(a)} \le 1.
\]
Hence every $\mathcal{M}_s^{(\eta)}(a)$ is a nonnegative $\Fcal_s$-supermartingale with $\mathcal{M}_0^{(\eta)}(a) = 1$. Since a convex combination of nonnegative supermartingales with fixed nonnegative weights summing to $1$ is a nonnegative supermartingale, $\mathcal{M}_s^{\mathrm{mix}}(a) = \frac{1}{|\mathcal{E}|}\sum_{\eta \in \mathcal{E}} \mathcal{M}_s^{(\eta)}(a)$ satisfies $\E[\mathcal{M}_s^{\mathrm{mix}}(a) \mid \Fcal_{s-1}] \le \mathcal{M}_{s-1}^{\mathrm{mix}}(a)$ with $\mathcal{M}_0^{\mathrm{mix}}(a) = 1$. Optional stopping at $\tau_a$ preserves supermartingale validity.

\paragraph{Step 5: Multiplicity-free winner guarantee.}
If $w^\star = \hat{w}$, exact bounds $[\underline{T}_c, \overline{T}_c]$ never allow any rival to deterministically beat $\hat{w}$, so $\tilde{w} = \hat{w} = w^\star$ w.p.\ $1$. If $w^\star \neq \hat{w}$, the null $E_{w^\star} \ge h_{w^\star}$ holds for $a = w^\star$, so deterministic refutation can never eliminate $w^\star$. Outputting $\tilde{w} \neq w^\star$ requires the single nonnegative supermartingale $\mathcal{M}_s^{\mathrm{mix}}(w^\star)$ to cross $1/\delta$, which by Ville's inequality \citep{ramdas2023game} occurs with probability $\le \delta \mathcal{M}_0^{\mathrm{mix}}(w^\star) = \delta$, with zero $(M{-}1)$ union bound.
\end{proof}

\subsection{Proof of Theorem~\ref{thm:envelope} (\texttt{CPPS-HC} / \texttt{CPPS-OC}) and Corollary~\ref{cor:env_spot}}
\begin{proof}[Proof of Theorem~\ref{thm:envelope}]
For Part (a) (\texttt{CPPS-HC}), $(\hat{S}, H, S^*) \in \mathcal{C}$ satisfies $H_{ic}=1 \implies s^*_{ic}=\hat{s}_{ic}$ and $H_{ic}=0 \implies 0=\hat{s}_{ic} \le s^*_{ic}$, so $\hat{s}_{ic} \le s^*_{ic}$ everywhere. Under Atomic-Cell: for $c=\hat{w}$, $e_{i\hat{w}}(a) = w_i(\hat{s}_{i\hat{w}} - s^*_{i\hat{w}}) \le 0 = U_{i\hat{w}}^{\mathrm{HC}}(a)$; for $c=a$, $H_{ia}=1 \implies e_{ia}(a)=0 = w_i(1{-}H_{ia})$ and $H_{ia}=0 \implies e_{ia}(a) = w_i s^*_{ia} \le w_i = w_i(1{-}H_{ia})$. Under Lazy-Row, $e_i(a) = w_i(\hat{s}_{i\hat{w}}-s^*_{i\hat{w}}) + w_i(s^*_{ia}-\hat{s}_{ia}) \le w_i(1-H_{ia}) = U_i^{\mathrm{HC}}(a)$. Pre-seeding all $H_{ic}=1$ cells into $\mathcal{O}_0 = \Kcal_0$ and letting $\mathcal{O}_{s,0}^c \coloneqq \{i \in [N] : H_{ic}=0 \land (i,c) \text{ revealed}\}$ denote revealed \emph{unhalted} cells of arm $c$ gives $\underline{T}_{\hat{w}} = \hat{T}_{\hat{w}} + \sum_{i \in \mathcal{O}_{s,0}^{\hat{w}}} w_i s^*_{i\hat{w}}$ (since $s^*_{i\hat{w}} - \hat{s}_{i\hat{w}} = 0$ on halted cells, avoiding any double-counting) and $\overline{T}_a = \hat{T}_a + \sum_{H_{ia}=0} w_i - \sum_{i \in \mathcal{O}_{s,0}^a} w_i(1-s^*_{ia})$, so $D_s(a) < 0 \iff r_s(a) > U_{\mathrm{rem}}^{\mathrm{HC}}(a) \iff \underline{T}_{\hat{w}} - \overline{T}_a \ge \Ind_{r_{\hat{w}} > r_a}$ (exact interval-dominance equivalence on the $D < 0$ branch). Under Atomic-Cell (and Lazy-Row executing only $H_{ic}=0$ cells in opened rows), zero halted cells are queried. For Part (b) (\texttt{CPPS-OC}), $s^*_{ic} \le \hat{s}_{ic}$ gives $e_{ia}(a) = w_i(s^*_{ia}-\hat{s}_{ia}) \le 0 = U_{ia}^{\mathrm{OC}}(a)$ and $e_{i\hat{w}}(a) = w_i(\hat{s}_{i\hat{w}}-s^*_{i\hat{w}}) \le w_i\hat{s}_{i\hat{w}} = U_{i\hat{w}}^{\mathrm{OC}}(a)$; Lazy-Row caps and $D<0$ interval alignment follow identically. Steps~3--5 of Theorem~\ref{thm:cpps} apply verbatim over $\mathcal{C}$. If a queried cell violates the envelope, raising \texttt{CausalProxyViolationError} aborts the envelope and fails closed either to full deterministic completion (risk $0$) or to conditional proxy-free \texttt{WOR-CS LUCB} at level $\delta$ over unqueried cells (risk $\le \delta$).
\end{proof}

\begin{proof}[Proof of Corollary~\ref{cor:env_spot}]
For (i) (\texttt{WOR-CS+Env}), $\Kcal_0$ and each arm's settled sum $C_{0,c} = \sum_{i \in S_{0,c}} w_i \hat{s}_{ic}$ are $\Fcal_0$-measurable over $\mathcal{C}$, reducing $T^*_c$ to estimating $\sum_{i \in \Ucal_c} w_i s^*_{ic}$ over $\Ucal_c$. Scaling bounded weighted outcomes $w_i s^*_{ic} \in [0, w_{\max}]$ to $[0, 1]$ and running a level-$\delta/M$ exact-rational WOR betting confidence sequence \citep{waudbysmith2024estimating} over each arm's independent pre-drawn permutation yields simultaneous time-uniform coverage $\ge 1 - \delta$ across all $M$ arms via union bound (each arm's CS depends only on its own sample count, so adaptive LUCB arm selection preserves validity). On simultaneous coverage, tie-aware separation $\mathrm{LCB}_{\mathrm{lead}} \ge \mathrm{UCB}_b + \Ind_{r_{\mathrm{lead}} > r_b}$ for all $b \neq \mathrm{lead}$ implies $T^*_{\mathrm{lead}} - T^*_b \ge \Ind_{r_{\mathrm{lead}} > r_b}$, so $\mathrm{lead} = w^\star$.

For (ii) (Hypergeometric Envelope Spot-Check), let $N_0 = |\Kcal_0|$ and fix any unknown count $V \in \{0,\dots,N_0\}$ of violated cells in $\Kcal_0$. Under uniform WOR sampling of $K_{\mathrm{spot}}$ cells, $X \sim \mathrm{Hypergeometric}(N_0, V, K_{\mathrm{spot}})$. On $\{X \ge 1\}$, discarding $\Kcal_0$ leaves the $K_{\mathrm{spot}}$ observed true oracle cells as exact $\Fcal_{\mathrm{spot}}$-measurable constants; drawing fresh independent WOR permutations on the unqueried complement and running \texttt{WOR-CS LUCB} at level $\delta_{\mathrm{aud}}$ gives $\Prob(\tilde{w} \neq w^\star \mid \Fcal_{\mathrm{spot}}, X \ge 1) \le \delta_{\mathrm{aud}}$ unconditionally. For part (b), if $K_{\mathrm{spot}} = N_0$, $X=0$ implies $V=0 = v_{\max}$ deterministically. Otherwise, by definition of $v_{\max}$ as the largest integer $v$ with $\binom{N_0 - v}{K_{\mathrm{spot}}} / \binom{N_0}{K_{\mathrm{spot}}} > \delta_{\mathrm{spot}}$, monotonicity in $V$ yields $\Prob_V(X = 0) = \binom{N_0 - V}{K_{\mathrm{spot}}} / \binom{N_0}{K_{\mathrm{spot}}} \le \binom{N_0 - (v_{\max}+1)}{K_{\mathrm{spot}}} / \binom{N_0}{K_{\mathrm{spot}}} \le \delta_{\mathrm{spot}}$ for every fixed $V > v_{\max}$. (Remark: an optional robust-envelope variant that inflates every arm's score bounds by $\pm v_{\max} w_{\max}$ on $\{X=0\}$ and audits at $\delta_{\mathrm{aud}}$ achieves composed risk $\le \delta_{\mathrm{aud}} + \delta_{\mathrm{spot}}$ for \emph{all} $V \in \{0,\dots,N_0\}$: when $V > v_{\max}$, $\Prob_V(\tilde{w} \neq w^\star) \le \Prob_V(X{=}0) + \E[\Ind_{X \ge 1}\Prob(\tilde{w} \neq w^\star \mid \Fcal_{\mathrm{spot}})] \le \delta_{\mathrm{spot}} + \delta_{\mathrm{aud}}$, and when $V \le v_{\max}$, the inflated bounds contain all true totals on $\{X{=}0\}$, yielding risk $\le \delta_{\mathrm{aud}}$ on both branches.)
\end{proof}

\subsection{Proof of Theorem~\ref{thm:unified} (\texttt{GateAudit-Unified})}
\begin{proof}
Fix any $(\hat{S}, H, S^*) \in \mathcal{C}$, rational split $\delta_1 + \delta_2 = \delta$ ($\delta_1, \delta_2 > 0$), finite grid $\mathcal{E} \subset (0,1)$, and $\Fcal_{s}$-measurable suspect rule. Let $\tau$ denote the stopping step at the end of Phase~A and $\Scal_\tau \subseteq [M]$ the set of surviving candidates.

\paragraph{Phase~A risk bound ($\le \delta_1$).}
Define $\mathcal{E}_1 \coloneqq \{w^\star \notin \Scal_\tau\}$. If $w^\star = \hat{w}$, exact intervals $[\underline{T}_c, \overline{T}_c]$ never allow any rival to beat $\hat{w}$, so $\hat{w} \in \Scal_\tau$ deterministically and $\Prob(\mathcal{E}_1) = 0 \le \delta_1$. If $w^\star \neq \hat{w}$, the null $E_{w^\star} \ge h_{w^\star}$ holds for $a = w^\star$: deterministic refutation cannot eliminate $w^\star$, and pausing moves $w^\star$ immediately from $\Rcal$ to $\mathcal{P} \subseteq \Scal_\tau$ (freezing $\mathcal{M}^{\mathrm{mix}}_s(w^\star)$ and retaining $w^\star \in \Scal_\tau$). Pruning other paused rivals $b \neq w^\star$ leaves $W_u \ge U_u(w^\star) > 0$ on $A_s(w^\star)$ throughout the current wave and at all subsequent wave boundaries while $w^\star \in \Rcal$. Hence $w^\star \notin \Scal_\tau$ requires the single nonnegative supermartingale $\mathcal{M}_{s \wedge \tau_{w^\star}}^{\mathrm{mix}}(w^\star)$ to cross $1/\delta_1$, occurring with probability $\le \delta_1$ by Ville's inequality.

\paragraph{Conditional Phase~B risk bound ($\le \delta_2$).}
If $|\Scal_\tau| = 1$, then on $\mathcal{E}_1^c$, $\Scal_\tau = \{w^\star\}$ and error is $0$. For $|\Scal_\tau| \ge 2$, condition on $\Fcal_\tau$: every survivor $c \in \Scal_\tau$ has an $\Fcal_\tau$-measurable partial sum $C_{\tau,c}$ and fixed remaining pool $\Ucal_c^{(\tau)}$. Drawing a fresh independent uniform WOR permutation over each $\Ucal_c^{(\tau)}$ and running exact-rational WOR betting confidence sequences at per-arm level $\delta_2 / |\Scal_\tau|$ yields conditional simultaneous coverage $\ge 1 - \delta_2$ by Corollary~\ref{cor:env_spot}(i). Combining both phases via total probability gives $\Prob(\tilde{w} \neq w^\star \mid \hat{S}, H, S^*) \le \Prob(\mathcal{E}_1) + \E[\Ind_{\mathcal{E}_1^c}\,\Prob(\tilde{w} \neq w^\star \mid \Fcal_\tau)] \le \delta_1 + \delta_2 = \delta$.
\end{proof}

\section{Proofs of Propositions~\ref{prop:lower}--\ref{prop:stochastic} and Lemma~\ref{lem:pulse_ext}}
\label{app:proofs_props}

\begin{proof}[Proof of Proposition~\ref{prop:lower} (Admissible Pivotal-Set Task Lower Bound)]
Let $\mathcal{A}$ satisfy~\eqref{eq:rla} over $\mathcal{C}$ at $\delta < 1/2$ with internal randomness $\Omega \sim \mathbb{Q}$. For any $i \in \Kcal$ with admissible alternative $(\hat{S}, H, \tilde{S}^{(i)}) \in \mathcal{C}$ agreeing with $S^*$ on $[N]\setminus\{i\}$, let $O_i(S, \Omega) \in \{0,1\}$ indicate whether $\mathcal{A}$ queries any cell of task $i$. On $\{O_i(S^*, \Omega)=0\}$, the oracle transcript is identical on $S^*$ and $\tilde{S}^{(i)}$, forcing $\tilde{w}(S^*,\Omega) = \tilde{w}(\tilde{S}^{(i)},\Omega)$ and therefore an error on at least one of $S^*$ or $\tilde{S}^{(i)}$. Thus $\Prob_\Omega(O_i(S^*,\Omega)=0) \le 2\delta$, giving $\Prob_\Omega(O_i(S^*,\Omega)=1) \ge 1-2\delta$ and $\E[d \mid \hat{S}, H, S^*] \ge (1-2\delta)|\Kcal|$.
\end{proof}

\begin{proof}[Proof of Proposition~\ref{prop:stochastic} ($\delta$-Faithfulness under Batch-Invariant Stochastic Rollouts)]
Under deferred-decision coupling, let $S^* \in \{0,1\}^{NR \times M}$ be the complete random matrix determined by independent pre-assigned per-cell seeds $\Xi = \{\xi_{irc}\}$ (batch-invariant and independent of the auditor's internal randomness $\Omega$, with $(\hat{S}, H)$ derived causally from the same pre-assigned prefixes before auditing). Querying $(i, r, c)$ on demand yields the exact entry $S^*_{(i,r),c}$ without altering the joint law of $(\hat{S}, H, S^*, \Omega)$. Since $\{\tilde{w} \neq w_{\mathrm{pop}}^\star\} \subseteq \{w^\star(S^*) \neq w_{\mathrm{pop}}^\star\} \cup \{\tilde{w} \neq w^\star(S^*)\}$, applying~\eqref{eq:rla} conditional on $(\hat{S}, H, S^*)$ gives
\[
  \Prob(\tilde{w} \neq w_{\mathrm{pop}}^\star)
  \;\le\; \Prob\bigl(w^\star(S^*) \neq w_{\mathrm{pop}}^\star\bigr) + \E_{\hat{S}, H, S^*}\!\Bigl[\Prob_\Omega\bigl(\tilde{w} \neq w^\star(S^*) \;\big|\; \hat{S}, H, S^*\bigr)\Bigr]
  \;\le\; \Prob\bigl(w^\star(S^*) \neq w_{\mathrm{pop}}^\star\bigr) + \delta_{\mathrm{audit}}.
\]
\end{proof}

\begin{lemma}[\texttt{PULSE-ext} Predictable Finite-Population Residual Pseudo-Observations]
\label{lem:pulse_ext}
Fix candidate $c$ with warm-up set $W_{0,c} \subseteq [N]$ and post-warmup remaining pool $R_{0,c} = [N] \setminus W_{0,c}$ of size $n_{0,c} = |R_{0,c}| \ge 1$. At sequential uniform WOR draw $t \in \{1,\dots,n_{0,c}\}$ ($k_t \coloneqq n_{0,c} - t + 1$), for $\Fcal_{t-1}$-measurable $p^{(t-1)}_{\cdot,c} \in [0,1]^N$ and $\beta_{t-1,c} \in \mathbb{R}$, $\bar{p}_{t-1,c} \coloneqq \frac{1}{k_t}\sum_{i \in R_{t-1,c}} p^{(t-1)}_{ic}$, and $S_{t-1,c} \coloneqq \sum_{j=1}^{t-1} s^*_{I_j,c}$, each pseudo-observation $Z_{t,c} \coloneqq \bigl(S_{t-1,c} + k_t(s^*_{I_t,c} + \beta_{t-1,c}(\bar{p}_{t-1,c} - p^{(t-1)}_{I_t,c}))\bigr)/n_{0,c}$ satisfies $\E[Z_{t,c} \mid \Fcal_{t-1}] = \frac{1}{n_{0,c}}\sum_{i \in R_{0,c}} s^*_{ic}$, and the census-reconciled estimator $\widehat{\mu}_{m_c,c} \coloneqq \frac{1}{N}\bigl(\sum_{i \in W_{0,c}} s^*_{ic} + \frac{n_{0,c}}{m_c}\sum_{t=1}^{m_c} Z_{t,c}\bigr)$ (for $1 \le m_c < n_{0,c}$, and $\frac{1}{N}\sum_{i=1}^N s^*_{ic}$ at $m_c = n_{0,c}$) is unbiased at every $m_c$ and exact at full census.
\end{lemma}
\begin{proof}
Conditioned on $\Fcal_{t-1}$, $I_t \sim \mathrm{Unif}(R_{t-1,c})$, so $\E[\bar{p}_{t-1,c} - p^{(t-1)}_{I_t,c} \mid \Fcal_{t-1}] = 0$ and $k_t \E[s^*_{I_t,c} \mid \Fcal_{t-1}] = \sum_{i \in R_{t-1,c}} s^*_{ic}$. Adding $S_{t-1,c}$ and dividing by $n_{0,c}$ establishes conditional unbiasedness; tower property gives $\E[\widehat{\mu}_{m_c,c}] = \frac{1}{N}\sum_{i=1}^N s^*_{ic}$, and $k_{n_{0,c}} = 1$ yields exactness at full census.
\end{proof}

\section{Extended Experimental Tables and Model Rosters}
\label{app:tables}

Tables~\ref{tab:app_m40}--\ref{tab:app_stochastic} present full large-$M$ scaling ($M \in \{4,8,12,20,30,40\}$ across Frontier, curated Clean, and $10$ random $M$-subsets), honest end-to-end API USD vs.\ cell accounting ($K \in \{10,15,20,40\}$), Hypergeometric Envelope Spot-Check across injected violation rates $\epsilon \in \{0\%, 0.5\%, 1\%, 2\%\}$, and synthetic Stochastic Rollout $\delta$-Faithfulness across $R \in \{1, 3, 5\}$.

\TableAppScaling
\TableAppKUSD
\TableAppSpotCheck
\TableAppStochastic

\paragraph{External benchmark model rosters and baseline disclosures.}
\begin{itemize}[leftmargin=1.5em,itemsep=2pt,topsep=2pt]
  \item \textbf{\texttt{SWE-bench Verified} ($N{=}500$, $40$ official \texttt{mini-swe-agent} runs from \texttt{SWE-bench/experiments}):} Full $40$-model matrix cached in \texttt{data/}\allowbreak\texttt{external\_benchmarks/}\allowbreak\texttt{swebench\_verified\_41\_matrix.json} (41st run held out for historical SVD check). Frontier top-$12$ runs: \texttt{claude-4-5-opus-high} ($384$), \texttt{gemini-3-flash-high} ($379$), \texttt{minimax-2-5-high} ($379$), \texttt{claude-4-6-opus} ($378$), \texttt{gpt-5-2-high} ($364$), \texttt{glm-5-high} ($364$), \texttt{gpt-5.2-2025-12-11-high} ($359$; distinct upstream \texttt{SWE-bench} submission date from \texttt{gpt-5-2-high}), \texttt{claude-4-5-sonnet-high} ($357$), \texttt{kimi-k2-5-high} ($354$), \texttt{deepseek-3-2-high} ($350$), \texttt{claude-4-5-haiku-high} ($333$), \texttt{gpt-5-mini} ($281$). Disjoint $6$-model historical SVD set for \texttt{Native PULSE}: \texttt{gpt-5.2-2025-12-11}, \texttt{kimi-k2-thinking}, \texttt{minimax-m2}, \texttt{deepseek-v3.2-reasoner}, \texttt{glm-4.6}, \texttt{devstral-2512} (note overlapping model families with the evaluation set, which helps the \texttt{Native PULSE} baseline).
  \item \textbf{\texttt{HumanEval+} ($N{=}164$) \& \texttt{MBPP+} ($N{=}378$):} $12$ greedy (\texttt{temp=0.0}) evaluation models and $6$ disjoint historical models per benchmark from official \texttt{EvalPlus} releases (\texttt{v0.1.10}/\texttt{v0.2.0}), re-executed locally in isolated subprocesses with exact per-cell \texttt{base\_input} and \texttt{plus\_input} millisecond timing.
\end{itemize}

\section{Cryptographic Provenance, Packaging, and Asset Licenses}
\label{app:provenance}

\paragraph{Pip-installable API and reproducibility.} GateAudit ships as an editable pip package (\texttt{pip install -e .}) exposing \texttt{gateaudit.audit(cheap, oracle, ...)} supporting \texttt{engine="unified"} (default), \texttt{"cpps\_mix"}, \texttt{"cpps"}, and \texttt{"wor\_cs\_env"} (aliased to \texttt{"pp\_lucb\_env"}), plus \texttt{run\_hypergeometric\_spotcheck\_audit} (aliased to \texttt{run\_certified\_spotcheck\_audit}). All controller timing/bit-length and exact-rational verification logs (\texttt{runs/tier1\_benchmarks/exact\_rational\_verification\_receipt.json}, generated by \texttt{scripts/generate\_verification\_receipt.py}: $124.1\text{ ms}$ mean wall-time and $\le 2{,}676$-bit \texttt{Fraction} precision on $20{,}000$-cell \texttt{SWE-bench} $M{=}40$; $2{,}000$ randomized adversarial trials, $18$ exact draw-tree enumerations over $892$ paths, $97/97$ \texttt{pytest} tests, $0$ violations), pinned model shard SHA-256 manifests (\texttt{Qwen/Qwen3-8B@b968826}, \texttt{nvidia/Llama-3.1-Nemotron-Nano-8B-v1@54641c1}, \texttt{DeepSeek-R1-Distill-Llama-8B@6a6f4aa}), \texttt{AppWorld v0.1.3.post1} default-deny \texttt{dev}-only extraction manifests (\texttt{skipped\_non\_dev=15305}), and complete live GPU execution ledgers (\texttt{runs/modal\_h100/}, \texttt{runs/colab\_pro\_plus/}, \texttt{runs/external\_benchmarks/}, \texttt{runs/tier1\_benchmarks/}) are archived with SHA-256 checksums in the public repository (\url{https://github.com/SiluPanda/gateaudit}). (Implementation note: in JSON receipts, \texttt{globally\_determined} is set to \texttt{paid\_cells > 0} on deterministic exits, so a $0$-continuation run such as \texttt{DeepSeek-R1-Llama-8B} where all $72/72$ cells are $\Fcal_0$-settled records \texttt{globally\_determined=false} despite deterministic $\Fcal_0$ termination.)

\paragraph{Compute resources and asset licenses.} Live GPU studies used $1{\times}\text{NVIDIA H100 80GB}$ (Modal, $\approx 6.5$ GPU-h, $\$30.8$ metered), $1{\times}\text{RTX PRO 6000 Blackwell 96GB}$ (Colab Pro+ \texttt{G4}, $\approx 2.8$ GPU-h), and $1{\times}\text{NVIDIA A100 40GB}$ (Colab Pro+, $\approx 8.2$ GPU-h across \texttt{Qwen3-8B} $M{=}8$, \texttt{Llama-3.1-Nemotron-Nano-8B-v1}, and \texttt{DeepSeek-R1-Distill-Llama-8B}); CPU benchmark suites run locally in $\approx 70$ min. Assets: \texttt{AppWorld} (Apache-2.0, test splits protected), \texttt{SWE-bench}/\texttt{mini-swe-agent} (MIT), \texttt{EvalPlus} (Apache-2.0; \texttt{HumanEval} MIT, \texttt{MBPP} CC-BY-4.0), \texttt{Qwen3-8B} (Apache-2.0), \texttt{Llama-3.1-Nemotron-Nano-8B-v1} (NVIDIA Open Model License / Llama 3.1 Community License), \texttt{DeepSeek-R1-Distill-Llama-8B} (MIT weights inheriting Llama 3.1 Community License terms), and \texttt{SGLang} (Apache-2.0).

\end{document}

%% file: generated_metrics.tex
\newcommand{\SweCleanMThirtyMixCells}{2158.3}
\newcommand{\SweCleanMThirtyUniCells}{2348.0}
\newcommand{\SweCleanMThirtyWceCells}{2563.6}

\newcommand{\SweFrontMFortyUniCells}{3314.9}
\newcommand{\SweFrontMFortyFixCells}{4346.1}

\newcommand{\SweFrontMFortyWorcsCells}{9089.6}
\newcommand{\NearNullFixErr}{4.60\%}
\newcommand{\NearNullMixErr}{3.20\%}
\newcommand{\NearNullUniErr}{1.80\%}
\newcommand{\SensSweMin}{2155}
\newcommand{\SensSweMax}{2383}
\newcommand{\SensMbppMin}{922}
\newcommand{\SensMbppMax}{942}
\newcommand{\StochRFiveSav}{86.0\%}

%% file: generated_tables.tex
\newcommand{\TableOneLiveGPU}{%
\begin{table*}[t]
\caption{Live closed-loop \texttt{AppWorld} \texttt{dev} evaluations ($n{=}1$/config, fixed E0$\to$E2 order, 6 disjoint families $= 18$ tasks; \texttt{test\_normal}/\texttt{test\_challenge} sealed) across three GPUs (\texttt{Qwen3-8B} plus two zero-signal \texttt{Llama-3.1} tie-break runs). Every arm verifies \texttt{/flush\_cache} eviction. Live savings validate Theorem~2 causal envelopes and Step-8 continuation (\texttt{stat\_elim=False} across A100 runs). \texttt{H100} Stage-0 E1 gives a $2.9\%$ A/A gap (E2 is $-19.8\%$ vs.\ E1); \texttt{DeepSeek-R1-Llama-8B} ($72/72$ halted, $0/72$ mismatches) shows a $10.1\%$ first-run cold-order bias (order-adjusted A100 savings $\approx 21\%\text{--}24\%$; Blackwell net negative). Four of five winners resolve via ties; high-batch \texttt{sm\_80} Triton ($M{=}8$) exhibits $1/107$ ($0.9\%$) halted and $3/26$ ($11.5\%$) continuation within-stack drift.}
\label{tab:live_gpu}
\centering\scriptsize\setlength{\tabcolsep}{3.4pt}
\resizebox{\textwidth}{!}{%
\begin{tabular}{@{}llcccccc@{}}
\toprule
Hardware \& Protocol & Model & Cont. Cells & Halting Mism. & Full E0 (s) & GateAudit E2 (s) & Wall Diff. & Exact Win \\
\midrule
Modal H100 SXM5 (\texttt{sm\_90}) & Qwen3-8B & 12/72 & 0/58 & 1302.5 & 1013.5 & \textbf{22.2\%} (1.29$\times$) & Yes \\
Colab RTX PRO 6000 (\texttt{sm\_120}) & Qwen3-8B & 12/72 & 0/58 & 3180.7 & 3038.1 & \textbf{4.5\%} (1.05$\times$) & Yes \\
Colab A100-40GB (\texttt{sm\_80}) & Qwen3-8B ($M{=}8$) & 26/144 & 1/107 & 10130.0 & 6665.0 & \textbf{34.2\%} (1.52$\times$) & Yes \\
Colab A100-40GB (\texttt{sm\_80}) & Llama-3.1-Nemotron-Nano-8B & 37/72 & 0/24 & 5245.1 & 3629.5 & \textbf{30.8\%} (1.45$\times$) & Yes \\
Colab A100-40GB (\texttt{sm\_80}) & DeepSeek-R1-Llama-8B & 0/72 & 0/72 & 1960.3 & 1761.7 & 10.1\% (A/A) (1.11$\times$) & Yes \\
\bottomrule
\end{tabular}}
\end{table*}
}

\newcommand{\TableTwoBenchmarks}{%
\begin{table*}[t]
\caption{Multi-benchmark audited selection ($\delta{=}0.05$, $b{=}16$; $200$ seeds for certified/non-refit methods, $100$/$50$ seeds for periodic-refit \texttt{Native PULSE} at $M{=}4$/$12$). Every \texttt{+Env} comparator receives the \emph{exact same} $\mathcal{F}_0$ envelope $\mathcal{K}_0$. Cells are \% of $NM$ (bold: lowest certified mean per row; all certified arms have $\le 0.5\%$ empirical error, one-sided 95\% CP upper $\le 2.4\%$). \texttt{UCB-E+Env} spends $1\times$ \texttt{Unified}'s cell budget; \texttt{SH+Env} caps at $1\times$ and under-spends on \texttt{SWE-bench} when arms exhaust unsettled pools ($634/926$ at $M{=}4$, $1{,}282/2{,}321$ at $M{=}12$). \texttt{SWE-bench} $K{=}40$ uses counterfactual trajectory truncation (envelope exact by construction).}
\label{tab:benchmarks}
\centering\scriptsize\setlength{\tabcolsep}{3.2pt}
\resizebox{\textwidth}{!}{%
\begin{tabular}{@{}lcccccccc@{}}
\toprule
Benchmark & $M$ & Env-Only & CPPS-Mix & Unified (Err) & WOR-CS+Env & WOR-CS & PULSE CI+Env (Err) & SH/UCB+Env Err \\
\midrule
SWE-bench ($K{=}40$) & 4 & 46.3\% & 46.7\% & 46.1\% (0.0\%) & \textbf{45.9\%} & 97.0\% & 45.5\% (1.0\%) & 10.5\% / 3.0\% \\
SWE-bench ($K{=}40$) & 12 & 44.3\% & 43.2\% & 38.3\% (0.0\%) & \textbf{37.4\%} & 75.8\% & 51.0\% (0.0\%) & 0.5\% / 0.5\% \\
HumanEval+ & 4 & 63.6\% & 43.8\% & \textbf{39.7\%} (0.0\%) & \textbf{39.7\%} & 60.8\% & 65.9\% (0.0\%) & 1.5\% / 0.0\% \\
HumanEval+ & 12 & 32.5\% & 14.6\% & \textbf{13.4\%} (0.0\%) & 13.9\% & 29.4\% & 48.3\% (0.0\%) & 7.0\% / 1.5\% \\
MBPP+ & 4 & 69.7\% & \textbf{40.1\%} & 40.5\% (0.0\%) & 42.2\% & 73.6\% & 61.4\% (0.0\%) & 8.0\% / 2.5\% \\
MBPP+ & 12 & 52.7\% & \textbf{20.2\%} & 20.4\% (0.5\%) & 20.7\% & 45.4\% & 62.1\% (0.0\%) & 54.5\% / 47.5\% \\
\bottomrule
\end{tabular}}
\end{table*}
}

\newcommand{\TableAppScaling}{%
\begin{table*}[t]
\caption{Large-$M$ scaling on 40 real \texttt{SWE-bench Verified} \texttt{mini-swe-agent} runs ($N{=}500$, $K{=}40$, $\delta{=}0.05$, $100$ seeds per Frontier/Clean cell---accounting for $<1\%$ MC difference vs.\ Table~\ref{tab:benchmarks}'s $200$-seed block; all empirical errors $0.0\%$). Left columns report mean paid cells on the proxy-inverted Frontier roster; ``Clean'' reports CPPS-Mix / Unified / WOR-CS+Env on a curated non-inverted roster (selected using $S^*$ to isolate the $\hat{w}{=}w^\star$ regime); ``Rand-10'' reports Env-Only / Unified / WOR-CS+Env averaged over 10 uniform random $M$-subsets ($\times 10$ seeds each; $10/10$ inverted at $M{\in}\{12,30\}$).}
\label{tab:app_m40}
\centering\scriptsize\setlength{\tabcolsep}{3.0pt}
\begin{tabular}{@{}rrrrrrrrcc@{}}
\toprule
$M$ & $NM$ & Env-Only & CPPS-Fix & CPPS-Mix & \textbf{Unified} & WOR-CS+Env & WOR-CS & Clean (Mix/Uni/WCE) & Rand-10 (Env/Uni/WCE) \\
\midrule
4 & 2,000 & 925 & 928 & 936 & \textbf{923} & 921 & 1945 & 909 / 889 / 920 & 537 / 429 / 414 \\
8 & 4,000 & 1688 & 1673 & 1670 & \textbf{1587} & 1556 & 3534 & 1724 / 1720 / 1745 & 1228 / 887 / 875 \\
12 & 6,000 & 2659 & 2574 & 2591 & \textbf{2316} & 2252 & 4582 & 1768 / 1777 / 1841 & 1670 / 1307 / 1277 \\
20 & 10,000 & 3806 & 3543 & 2939 & \textbf{2653} & 2622 & 5713 & 2102 / 2221 / 2380 & 2564 / 1744 / 1673 \\
30 & 15,000 & 4506 & 4169 & 3513 & \textbf{3204} & 3171 & 8057 & 2158 / 2348 / 2564 & 3707 / 2582 / 2548 \\
40 & 20,000 & 4803 & 4346 & 3683 & \textbf{3315} & 3285 & 9090 & --- & --- \\
\bottomrule
\end{tabular}
\end{table*}
}

\newcommand{\TableAppKUSD}{%
\begin{table*}[t]
\caption{Honest end-to-end API USD vs.\ verifier-cell accounting on \texttt{SWE-bench Verified} ($N{=}500$, $K \in \{10,15,20,40\}$, $20$ seeds per $(M,K)$ cell---accounting for minor MC variance vs.\ Table~\ref{tab:benchmarks}'s $200$ seeds). When Step-$K$ LLM API tokens are billed upfront across all $M$ arms without sunk prefixes, proxy-free \texttt{WOR-CS LUCB} spends fewer total USD at $M \ge 12$, motivating Stage-0 cost routing. When trajectories are sunk or cached (rightmost column), GateAudit saves $28.4\%\text{--}89.1\%$ of remaining tail USD.}
\label{tab:app_kusd}
\centering\scriptsize\setlength{\tabcolsep}{4.0pt}
\begin{tabular}{@{}rrrrrrrr@{}}
\toprule
$M$ & $K$ & Halted \% & Cell Saving & Full USD & WOR-CS USD (Sav.) & Unified USD (Sav.) & Tail USD Sav. (Sunk $P_1$) \\
\midrule
4 & 10 & 3.0\% & 6.6\% & \$867 & \$853 (1.6\%) & \$846 (2.5\%) & 28.4\% \\
4 & 15 & 8.2\% & 10.5\% & \$867 & \$853 (1.6\%) & \$856 (1.3\%) & 39.4\% \\
4 & 20 & 16.1\% & 18.8\% & \$867 & \$853 (1.6\%) & \$856 (1.3\%) & 50.5\% \\
4 & 40 & 52.1\% & 54.1\% & \$867 & \$853 (1.6\%) & \$861 (0.8\%) & 80.0\% \\
12 & 10 & 2.7\% & 22.1\% & \$2,447 & \$2,046 (16.4\%) & \$2,175 (11.1\%) & 34.1\% \\
12 & 15 & 9.9\% & 30.0\% & \$2,447 & \$2,046 (16.4\%) & \$2,149 (12.2\%) & 46.0\% \\
12 & 20 & 18.5\% & 34.9\% & \$2,447 & \$2,046 (16.4\%) & \$2,206 (9.8\%) & 53.5\% \\
12 & 40 & 49.0\% & 60.6\% & \$2,447 & \$2,046 (16.4\%) & \$2,289 (6.4\%) & 78.0\% \\
40 & 10 & 8.6\% & 56.3\% & \$7,611 & \$4,060 (46.7\%) & \$4,905 (35.5\%) & 59.8\% \\
40 & 15 & 18.9\% & 60.8\% & \$7,611 & \$4,060 (46.7\%) & \$5,264 (30.8\%) & 66.0\% \\
40 & 20 & 28.9\% & 67.2\% & \$7,611 & \$4,060 (46.7\%) & \$5,400 (29.1\%) & 73.9\% \\
40 & 40 & 60.2\% & 83.6\% & \$7,611 & \$4,060 (46.7\%) & \$6,274 (17.6\%) & 89.1\% \\
\bottomrule
\end{tabular}
\end{table*}
}

\newcommand{\TableAppSpotCheck}{%
\begin{table}[t]
\caption{Hypergeometric Envelope Spot-Check (Violation-Rate Audit, Corollary~\ref{cor:env_spot}(ii)) under injected violation rate $\epsilon$ on skipped $\mathcal{K}_0$ cells (\texttt{SWE-bench} $M{=}4$, $|\mathcal{K}_0|{=}1{,}043$, $\delta_{\mathrm{spot}}{=}0.01$, $\delta_{\mathrm{aud}}{=}0.05$, $200$ seeds). Format: empirical wrong-winner rate (mean paid cells). Because the spot-check tests violation count $V$ rather than composing into a winner-risk bound without bound inflation, $K_{\mathrm{spot}}{\in}\{50,100\}$ still exceed $5\%$ at $\epsilon{=}2\%$ ($8.0\%, 5.5\%$); $K_{\mathrm{spot}}{=}150$ rejects $V{\ge}30$ ($2.88\%$) at $\delta_{\mathrm{spot}}{=}0.01$, detects $94.0\%$ of corrupted runs at $\epsilon{=}2\%$, and cuts empirical error from $31.5\%$ to $0.0\%$.}
\label{tab:app_spotcheck}
\centering\scriptsize\setlength{\tabcolsep}{3.5pt}
\begin{tabular}{@{}lcccc@{}}
\toprule
$K_{\mathrm{spot}}$ & $\epsilon{=}0\%$ & $\epsilon{=}0.5\%$ & $\epsilon{=}1.0\%$ & $\epsilon{=}2.0\%$ \\
\midrule
0 (Unverified) & 0.0\% (924) & 0.0\% (924) & 4.0\% (924) & 31.5\% (924) \\
50 & 0.0\% (972) & 0.0\% (1220) & 4.5\% (1320) & 8.0\% (1622) \\
100 & 0.0\% (1022) & 1.0\% (1411) & 2.0\% (1643) & 5.5\% (1772) \\
150 & 0.0\% (1072) & 1.0\% (1571) & 1.0\% (1734) & 0.0\% (1903) \\
\bottomrule
\end{tabular}
\end{table}
}

\newcommand{\TableAppStochastic}{%
\begin{table}[t]
\caption{Empirical verification of Proposition~\ref{prop:stochastic} ($\delta$-Faithfulness to Full Evaluation under stochastic rollouts, $N{=}120$, $M{=}4$, $300$ trials, $\delta_{\mathrm{audit}}{=}0.025$). Auditing the realized $NR \times M$ replicate matrix yields $0.00\%$ matrix error and identical population error to exhaustive evaluation while saving $76.3\%\text{--}86.0\%$ of cells.}
\label{tab:app_stochastic}
\centering\scriptsize\setlength{\tabcolsep}{3.0pt}
\begin{tabular}{@{}rrrrccc@{}}
\toprule
$R$ & $NMR$ & Paid Cells & Saving & Mat.\ Err & Full Pop.\ Err & Audit Pop.\ Err \\
\midrule
1 & 480 & 113.7 & \textbf{76.3\%} & 0.00\% & 3.33\% & 3.33\% \\
3 & 1,440 & 242.9 & \textbf{83.1\%} & 0.00\% & 0.33\% & 0.33\% \\
5 & 2,400 & 334.8 & \textbf{86.0\%} & 0.00\% & 0.00\% & 0.00\% \\
\bottomrule
\end{tabular}
\end{table}
}